\documentclass[sigconf]{acmart}
\AtBeginDocument{%
  }

\copyrightyear{2025}
\acmYear{2025}
\setcopyright{cc}
\setcctype{by}
\acmConference[MM '25]{Proceedings of the 33rd ACM International Conference on Multimedia}{October 27--31, 2025}{Dublin, Ireland}
\acmBooktitle{Proceedings of the 33rd ACM International Conference on Multimedia (MM '25), October 27--31, 2025, Dublin, Ireland}
\acmDOI{10.1145/3746027.3758152}
\acmISBN{979-8-4007-2035-2/2025/10}

\usepackage{graphicx}
\usepackage[linesnumbered,ruled]{algorithm2e}
\usepackage{multirow}
\usepackage{pifont}
\usepackage[dvipsnames]{xcolor}
\usepackage{colortbl}
\usepackage{xspace}
\newcommand{\etal}{\textit{et al.}\xspace}
\newcommand{\eg}{\textit{e.g.}\xspace}
\newcommand{\ie}{\textit{i.e.}\xspace}

\usepackage{enumitem}
\AtEndPreamble{
    \usepackage[capitalize]{cleveref}
    \crefname{section}{Sec.}{Secs.}
    \Crefname{section}{Section}{Sections}
    \crefname{table}{Tab.}{Tabs.}
    \Crefname{table}{Table}{Tables}
}
\usepackage{amsmath}
\usepackage{booktabs}
\usepackage{subcaption}

\begin{document}

\title[F3: A Training-free and Efficient Visual Adversarial Example Purification Method in LVLMs]{Fighting Fire with Fire (F3): A Training-free and Efficient Visual Adversarial Example Purification Method in LVLMs}

\author{Yudong Zhang}
\orcid{0009-0009-6049-603X}

\affiliation{%
  \institution{Tsinghua University, Tencent}
  \city{Beijing}
  \country{China}
}
\email{zhangyd16@mails.tsinghua.edu.cn}

\author{Ruobing Xie}
\orcid{0000-0003-3170-5647}
\authornote{Corresponding authors.}
\affiliation{%
  \institution{Tencent}
  \city{Beijing}
  \country{China}}
\email{xrbsnowing@163.com}

\author{Yiqing Huang}
\orcid{0000-0002-2143-3329}
\affiliation{%
 \institution{Tencent}
 \city{Beijing}
 \country{China}}
\email{huang-yq17@tsinghua.org.cn}

\author{Jiansheng Chen}
\orcid{0000-0002-2040-7938}
\authornotemark[1]
\affiliation{%
  \institution{University of Science and Technology Beijing}
  \city{Beijing}
  \country{China}
}
\email{jschen@ustb.edu.cn}

\author{Xingwu Sun}
\orcid{0009-0008-3222-0901}
\affiliation{%
 \institution{Tencent, University of Macau}
 \city{Beijing}
 \country{China}}
\email{sunxingwu01@gmail.com}

\author{Zhanhui Kang}
\orcid{0009-0006-5151-4222}
\affiliation{%
 \institution{Tencent}
 \city{Shenzhen, Guangdong}
 \country{China}}
\email{kegokang@tencent.com}

\author{Di Wang}
\orcid{0009-0003-6713-1638}
\affiliation{%
 \institution{Tencent}
 \city{Beijing}
 \country{China}}
\email{diwang@tencent.com}

\author{Yu Wang}
\orcid{0000-0001-6108-5157}
\authornotemark[1]
\affiliation{%
  \institution{Tsinghua University}
  \city{Beijing}
  \country{China}}
\email{yu-wang@mail.tsinghua.edu.cn}

\renewcommand{\shortauthors}{Yudong Zhang et al.}

\begin{abstract}
Recent advances in large vision-language models (LVLMs) have showcased their remarkable capabilities across a wide range of multimodal vision-language tasks. However, these models remain vulnerable to visual adversarial attacks, which can substantially compromise their performance. In this paper, we introduce F3, a novel adversarial purification framework that employs a counterintuitive ``fighting fire with fire'' strategy: intentionally introducing simple perturbations to adversarial examples to mitigate their harmful effects. Specifically, F3 leverages cross-modal attentions derived from randomly perturbed adversary examples as reference targets. By injecting noise into these adversarial examples, F3 effectively refines their attention, resulting in cleaner and more reliable model outputs. Remarkably, this seemingly paradoxical approach of employing noise to counteract adversarial attacks yields impressive purification results. Furthermore, F3 offers several distinct advantages: it is training-free and straightforward to implement, and exhibits significant computational efficiency improvements compared to existing purification methods. These attributes render F3 particularly suitable for large-scale industrial applications where both robust performance and operational efficiency are critical priorities. The code is available at \url{https://github.com/btzyd/F3}.
\end{abstract}





\maketitle

\section{Introduction}
\label{sec:intro}

Large vision-language models (LVLMs) have garnered significant attention for their remarkable multimodal capabilities in various applications, including but not limited to image classification, image captioning, and visual question answering (VQA) \cite{alayrac2022flamingo, liu2024visual, li2023blip, dai2024instructblip, zhu2023minigpt, zong2024mova, shi2024eagle, liu2024improved, bai2023qwenvlversatilevisionlanguagemodel, wang2024qwen2, bai2025qwen2}. Despite their robust performance in these tasks, such models are presented with notable challenges in the realm of adversarial robustness and vulnerability concerns.

\begin{figure}[!htbp]
  \centering
  \includegraphics[width=0.9\linewidth]{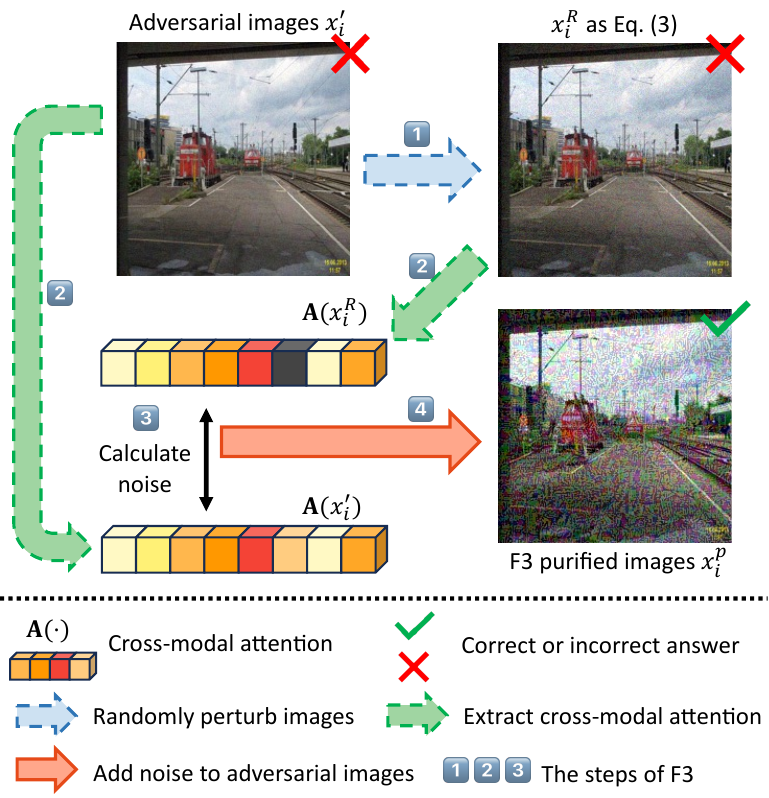}
  \caption{Overview of F3. (1) We inject random noise into adversarial examples $x_i'$ to generate perturbed images $x_i^R$. (2) Extracted cross-modal attention $\mathbf{A}(x_i^R)$ serves as the reference attention for purification purposes. (3) Using this reference attention as a target, we calculate the purification noise accordingly. (4) Surprisingly, by intentionally introducing noise to the adversarial examples, F3 effectively steers them toward alignment with the reference attention $\mathbf{A}(x_i^R)$, ultimately leading to more robust model responses.}
  \label{fig:motivation}
\end{figure}

Recent years have witnessed increasing research attention focused on adversarial attacks targeting Large Language Models (LVLMs) \cite{yin2023vlattack, dong2023robust, carlini2024aligned, schlarmann2023adversarial, zhang2022towards, zhao2024evaluating}. Due to the inherent high dimensionality and redundancy of visual data, combined with the architectural complexity of LVLMs, adversaries can readily manipulate these models into generating incorrect or even harmful responses by introducing carefully crafted adversarial perturbations to input images. This vulnerability poses significant risks to the robustness and reliability of LVLMs, particularly in high-stakes real-world applications. As a result, there is an urgent need for technical solutions to mitigate the vulnerability of LVLMs to adversarial examples.

Despite this critical challenge, research specifically addressing \emph{\textbf{adversarial example purification in LVLM}} remains limited. Existing adversarial purification methods are primarily designed for visual models rather than LVLMs, focusing on image-centric techniques such as random resizing and padding \cite{xie2018mitigating}, as well as compression \cite{jia2019comdefend} of adversarial examples. Some approaches leverage image generative models, including diffusion-based methods \cite{nie2022DiffPure}. However, these methods predominantly focus on the visual modality and fail to fully account for the unique multimodal interaction characteristics inherent in LVLMs. Consequently, they often exhibit non-robust purification performance for LVLMs or incur substantial computational costs, rendering them impractical for efficient deployment in real-world applications.

To address this challenge, we propose designing an training-free and efficient purification method specifically tailored for LVLMs by deeply exploring their central multimodal interactions. In a typical LVLM \cite{dai2024instructblip, zhu2023minigpt, liu2024visual, wang2024qwen2}, the visual encoder and vision-language projector process images into a sequence of visual tokens, aligning them with the input space of a large language model (LLM). These visual tokens are then combined with text tokens and other inputs before being fed collectively into the LLM to generate outputs. During the answer generation process within the LLM, newly generated tokens compute attention with image tokens and other tokens, integrating this information based on their attention weights. We define \emph{the attention of the text token (generated by the LLM) towards the visual tokens} as \textbf{cross-modal attention} $\mathbf{A}$, which serves as a key indicator of multimodal interaction in LVLMs. Our analysis reveals \textbf{significant differences in attention patterns between clean and adversarial examples}. This observation suggests that cross-modal attention may represent a critical vulnerability in LVLMs, inspiring us to explore novel adversarial purification strategies centered on this unique aspect of multimodal interaction. 

Based on the distinct differences in cross-modal attention between clean and adversarial examples, we propose an intuitive and bold idea: if it is possible to realign the attention of adversarial examples with those of their clean counterparts, could such alignment serve as an effective means for purifying these adversarial examples? However, this approach raises two critical challenges. First, \emph{\textbf{how do we determine the purification direction?}} Without access to the ground truth clean example during purification, establishing a reliable \emph{reference attention} becomes essential to guide our purification process. Second, \emph{\textbf{how do we modify the adversarial image to achieve purification?}} Once the reference attention is established, we must determine how to perturb adversarial images such that their cross-modal attention aligns with the reference attention.

To address these challenges, we introduce our novel, straightforward, effective, and efficient method, F3, as illustrated in \cref{fig:motivation}. Our approach consists of the following steps: (1) We first apply random perturbation to an adversarial example $x_i'$, resulting in $x_i^R$. (2) Next, we extract cross-modal attention from both $x_i'$ and $x_i^R$, denoted as $\mathbf{A}(x_i')$ and $\mathbf{A}(x_i^R)$ respectively. Importantly, the attention of the perturbed adversarial example $\mathbf{A}(x_i^R)$ serves as a rough estimate of the purification direction toward the ideal but inaccessible clean attention. (3) We compute the similarity between the attention of the original adversarial examples and their corresponding reference attentions to estimate the purification noise. (4) Finally, we apply this estimated noise directly to the original adversarial example. Surprisingly, despite adding further perturbations to adversarial examples (resulting in purified images $x_i^p$ that appear noisier than the original adversarial examples $x_i'$), F3 demonstrates remarkable purification effectiveness by aligning the attention more closely with clean attention. Extensive experiments validate both the effectiveness and efficiency of our approach. This approach of \emph{counteracting the perturbations of adversarial examples through simple and counterintuitive perturbations} is akin to \textbf{\underline{F}ighting \underline{F}ire with \underline{F}ire (F3)}. Unlike conventional purification approaches that primarily aim for visually pleasing outcomes, F3 adopts a distinct strategy by prioritizing the mitigation of vulnerabilities in LVLMs against adversarial attacks. It effectively and efficiently purifies adversarial examples while enabling LVLMs to maintain accurate output generation even when processing such adversarial examples.

Our contributions can be summarized as follows: 

\noindent
(1) We introduce F3, a novel and training-free adversarial purification method designed specifically for countering visual adversarial attacks on LVLM. It bravely and effectively purifies adversarial examples by incorporating additional noise guided through cross-modal attention to direct the purification process. Notably, F3 represents the \textbf{\emph{first dedicated adversarial purification approach via adding noise tailored to visual adversarial attacks in LVLMs}}. 

\noindent
(2) Our method leverages an innovative approach by introducing random perturbations to adversarial examples for estimating the direction of clean attention. This estimated direction serves as a reference guide for F3's additive noise generation, yielding a method that is \textbf{\emph{novel}}, \textbf{\emph{computationally efficient}}, and \textbf{\emph{training-free}}. 

\noindent
(3) Comprehensive empirical evaluations confirm the effectiveness of F3 across popular LVLMs such as BLIP-2 \cite{li2023blip}, InstructBLIP \cite{dai2024instructblip}, LLaVA-v1.5 \cite{liu2024improved}, and Qwen2.5-VL \cite{bai2025qwen2}. F3 successfully counters both non-adaptive and adaptive adversarial attacks under diverse scenarios, showcasing its strong potential for \textbf{\emph{real-world applications with minimal computational overhead}}. 

\section{Related Works}
\label{sec:related_works}

\subsection{Large Vision-Language Models (LVLMs)}
Large vision-language models are built on powerful vision encoders \cite{dosovitskiy2020image, fang2023eva, zhang2025enhancing} and large language models \cite{chiang2023vicuna, chung2024scaling, touvron2023llama, zhang2022opt}, with the two components integrated through a vision-language projector. The process involves extracting features from input images using the vision encoder, which are then encoded by the vision-language projector \cite{zhu2025connector, zhang2025security} into tokens compatible with the large language model's input space. These image tokens, combined with text tokens, are processed by the large language model to generate responses. Widely used vision-language projectors include Q-Former \cite{li2023blip, dai2024instructblip, zhu2023minigpt} and multi-layer perceptron (MLP) \cite{liu2024visual, shi2024eagle, wang2024qwen2, bai2023qwenvlversatilevisionlanguagemodel, bai2025qwen2}. 

\subsection{Adversarial Attacks and Defense Mechanisms}
Adversarial examples are generated by strategically introducing small, carefully crafted perturbations to inputs, causing models to produce erroneous outputs. Early studies on adversarial attacks focused primarily on unimodal vision models, employing methods such as FGSM \cite{goodfellow2014explaining}, PGD \cite{madry2017towards}, JSMA \cite{papernot2016limitations}, DeepFool \cite{moosavi2016deepfool}, and the Carlini \& Wagner (C\&W) attack \cite{carlini2017towards}. However, recent research has demonstrated that LVLMs are equally vulnerable to such attacks. For example, Attack-Bard \cite{dong2023robust} generates adversarial examples across multiple surrogate models, successfully targeting ChatGPT-4 and Bard. Similarly, Carlini \etal \cite{carlini2024aligned} utilized visual adversarial examples to trick LVLMs into producing harmful statements. In addition, QAVA \cite{zhang2025qava} is a query-agnostic visual attack method. To mitigate these threats, various defense mechanisms have been explored. Adversarial examples can be partially neutralized through techniques such as random resizing and padding \cite{xie2018mitigating}, image super-resolution \cite{mustafa2019image}, and image compression \cite{jia2019comdefend}, among others \cite{kang2021stable, ho2022disco}. Some studies have focused on detecting adversarial examples without purification \cite{zhang2024pip}, while other research has explored the use of generative models for adversarial example purification. For instance, PixelDefend \cite{song2018pixeldefend} leverages PixelCNN \cite{van2016conditional}, Defense-GAN employs GAN-based architectures \cite{samangouei2018defensegan, NIPS2014_5ca3e9b1}, and DiffPure utilizes diffusion models \cite{nie2022DiffPure, ho2020denoising}. However, these methods often apply generic filters or rely on separate models for image processing, without considering the specific inference model used. This limitation can lead to suboptimal robustness and high computational costs associated with training purification models and performing LVLM inference. In contrast, our work bridges this critical gap by specifically investigating adversarial purification within the context of LVLMs. We present a training-free, efficient, and model-agnostic approach that achieves this through deliberately introducing noise into adversarial examples during inference. 

\section{F3: Fighting Fire with Fire}
\label{sec:methods}

We first introduce our F3's motivation and method. Without loss of generality, we adopt the following experimental settings as a representative example for our investigation of F3 in \cref{sec:methods}. Comprehensive experiments in various configurations are given in \cref{sec:experiment}.

\noindent
\textbf{Datasets}. We utilize the $\mathcal{D}_\text{VQAv2}^{1000}$ dataset, which comprises 1,000 image-text pairs sampled from VQA v2 \cite{goyal2017making}. This dataset serves as a foundational benchmark for our preliminary experiments.

\noindent
\textbf{Models}. Unless otherwise specified, the InstructBLIP Vicuna-7B model is employed as the experimental LVLM in this section.

\noindent
\textbf{Attacks settings}. In this section, we focus on non-adaptive attacks to thoroughly elucidate the design principles and properties of our approach. Results under adaptive attack scenarios are presented in \cref{tab:compare_diff_models_adaptive}. For non-adaptive attacks, we primarily employ the Carlini \& Wagner (C\&W) method \cite{carlini2017towards}. The attack iterates for 50 steps with a step size of $0.01$ and sets the constant $c=0.005$, as defined in \cref{eq:cw_loss}, where $\mathcal{L}_\text{LVLM}(x_i', x_t)$ represents the loss of the LVLM when processing the adversarial image $x_i'$ and text $x_t$.
\begin{equation}
    \label{eq:cw_loss}
    \mathcal{L}_\text{C\&W}(x_i, x_i', x_t) = \mathcal{L}_\text{LVLM}(x_i', x_t) - c\times||x_i-x_i'||_2.
\end{equation}

\subsection{Cross-Modal Attention in LVLMs}
\label{sec:definition_of_attention}

We begin by reviewing the standard architecture of LVLMs. For a given input image $x_i$, the visual encoder $f_v$ extracts visual features, which are then processed by the vision-language projector to produce $M$ visual tokens $I = \{I_j | 1 \leq j \leq M\}$. Simultaneously, the tokenizer of the large language model (LLM) encodes the input text $x_t$ into $N$ text tokens $T = \{T_j | 1 \leq j \leq N\}$. These visual and text tokens are then jointly fed into the LLM for output generation. In total, the LLM processes $M + N$ input tokens $\{I, T\}$, computing self-attention across all tokens to produce sequential outputs. 

Building on insights from PIP \cite{zhang2024pip} and DHCP \cite{zhang2024dhcp}, we focus specifically on the attention patterns within the LLM during the generation of the first token. In decoder-only LLMs, when generating this initial token, the model calculates attention weights between the token and all preceding $M + N$ tokens, using these weights to aggregate information \cite{vaswani2017attention}. Our analysis centers on the cross-modal attention between the first response token and the visual tokens $I$, which we define as $\mathbf{A}(x_i, x_t, f)$ (abbreviated as $\mathbf{A}(x_i)$). This attention tensor typically has dimensions $(L, H, M)$, where $f$ represents the LVLM, $L$ is the number of layers in the LLM, and $H$ is the number of attention heads. Cross-modal attention plays a critical role in extracting visual information for multimodal tasks during response generation, making it particularly relevant to studies on visual adversarial attacks and purification.

\subsection{Cross-Modal Attention Differs Between Clean and Adversarial Examples}
\label{sec:diff_attention_clean_adv}

To investigate whether cross-modal attention $\mathbf{A}$ differs between clean and adversarial examples, we generated 1000 adversarial examples using C\&W untargeted attacks on the $\mathcal{D}_\text{VQAv2}^{1000}$ dataset. We evaluated the model's response performance on both clean and adversarial examples using VQA scores. The VQA score for clean examples was found to be 75.95, while for adversarial examples, it dropped significantly to 24.88.

We then investigated the impact of adversarial attacks on cross-modal attention $\mathbf{A}$, as defined in \cref{sec:definition_of_attention}. Our analysis revealed significant disparities between clean and adversarial examples in terms of $\mathbf{A}$. To facilitate visual comparison, we applied a maxima operation across the multi-head attention dimensions to project $\mathbf{A}$ into a two-dimensional representation. As illustrated in \cref{fig:pipeline}, there are notable differences between the attention of clean examples ($\mathbf{A}(x_i)$) and adversarial examples ($\mathbf{A}(x_i')$). Specifically, we observed an increase in attention values for the 8th token following the attacks. Similar patterns were consistently observed across various attack methods and different question types and number-related questions, with additional details provided in Appendix. The differences in cross-modal attentions between clean and adversarial examples are consistent and can be quantitatively assessed using metrics such as mean-square error (MSE) and Kullback-Leibler (KL) divergence, as presented in the cells where column ``Adv'' intersects with rows ``MSE'' and ``KL'' of \cref{tab:score_of_clean_ref}. These findings suggest potential strategies for mitigating or purifying adversarial examples.

\begin{table}[!htbp]
  \caption{The VQA scores and attention similarity for purified examples as defined in \cref{eq:clean_loss}, where $\mathbf{A}_\text{clean} = \mathbf{A}(x_i)$.}
  \label{tab:score_of_clean_ref}
  \centering
  \resizebox{\linewidth}{!}{
  \begin{tabular}{c|c|c|ccccc}
    \toprule
    \multirow{2}{*}{} & \multirow{2}{*}{Clean} & \multirow{2}{*}{Adv} & \multicolumn{5}{c}{The value of $\gamma_\infty$ as \cref{eq:clean_loss}} \\
    \cmidrule(lr){4-8}
    & & & 2/255 & 4/255 & 8/255 & 16/255 & 32/255  \\
    \midrule
    VQA scores $\uparrow$ & 75.95 & 24.88 & 27.11 & 27.85 & 30.65 & 36.77 & 45.89 \\
    MSE($\mathbf{A}_\text{clean}, \mathbf{A}$) $\downarrow$ & 0 & 10.31 & 12.34 & 11.74 & 10.85 & 9.65 & 7.89 \\
    KL($\mathbf{A}_\text{clean}, \mathbf{A}$) $\downarrow$ & 0 & 3.39 & 3.79 & 3.63 & 3.41 & 3.04 & 2.54 \\
    \bottomrule
  \end{tabular}
  }
\end{table}

\subsection{Clean Example's Attention is a Good Goal for Adversarial Purification}
\label{sec:clean_attention}

As demonstrated in \cref{sec:diff_attention_clean_adv}, there exists a significant difference in the cross-modal attention $\mathbf{A}$ between clean and adversarial examples. This observation raises an important question: Can adversarial example purification be achieved by optimizing their attention to more closely resemble that of clean examples? To address this, we propose aligning the attention of adversarial examples with that of clean examples through a process of deliberately introducing noise to the adversarial examples, thereby reducing their attention difference. This approach is formally described in \cref{eq:clean_loss}, where $x_i^p$ represents the purified version of the adversarial example $x_i'$.
\begin{align}
    \label{eq:clean_loss}
        x_i^p = x_i' - \gamma\times\text{sign}(\nabla_{x_i'}||&\mathbf{A}(x_i', x_t, f)- \mathbf{A}(x_i, x_t, f)||_2), \\
        \gamma&\sim\mathcal{U}[0, \gamma_\infty]. \notag
\end{align}
In our implementation, we selected $\gamma$ as a series of values bounded by $\gamma_\infty$ and applied perturbations to the adversarial examples according to \cref{eq:clean_loss}. The results are summarized in \cref{tab:score_of_clean_ref}. Our experiments reveal that when the optimization objective is focused \textbf{solely on aligning the attention of the adversarial example with that of the clean example, the adversarial example undergoes significant purification}, as evidenced by the improvement in the VQA score. Furthermore, the optimization step $\gamma$ exhibits an increasing trend with higher $\gamma_\infty$, leading to a corresponding improvement in both the VQA score and attention similarity metrics.

We also assessed the similarity between the attention patterns of clean, adversarial, and purified examples using two evaluation metrics: Mean Squared Error (MSE) and Kullback-Leibler (KL) divergence. Specifically, for MSE, we calculated the squared differences between pairs of attention sets and then averaged these values across all comparisons. For KL divergence, we measured the disparity between the two attention probability distributions by first normalizing the attention weights across the $M$ visual tokens within each attention head of every layer, effectively treating them as valid probability distributions. Our analysis revealed that, according to both MSE and KL metrics, the attention patterns of purified examples exhibited a significantly higher similarity to those of clean examples compared to adversarial examples. In this study, we employed clean attention that is inherently inaccessible during the purification process. This approach was specifically chosen given that our primary goal was to investigate the feasibility of optimizing adversarial example attention to facilitate their alignment with clean counterparts for effective purification. \textbf{\Cref{tab:score_of_clean_ref} demonstrate that the attention alignment method successfully purifies adversarial examples when clean attention is available}.

\subsection{F3-v1: Cross-Modal Attention of Randomly-Perturbed Adversarial Example as a Practical Reference for Purification}
\label{sec:random_noise_attention}

\begin{figure*}[!htbp]
  \centering
  \includegraphics[width=0.9\linewidth]{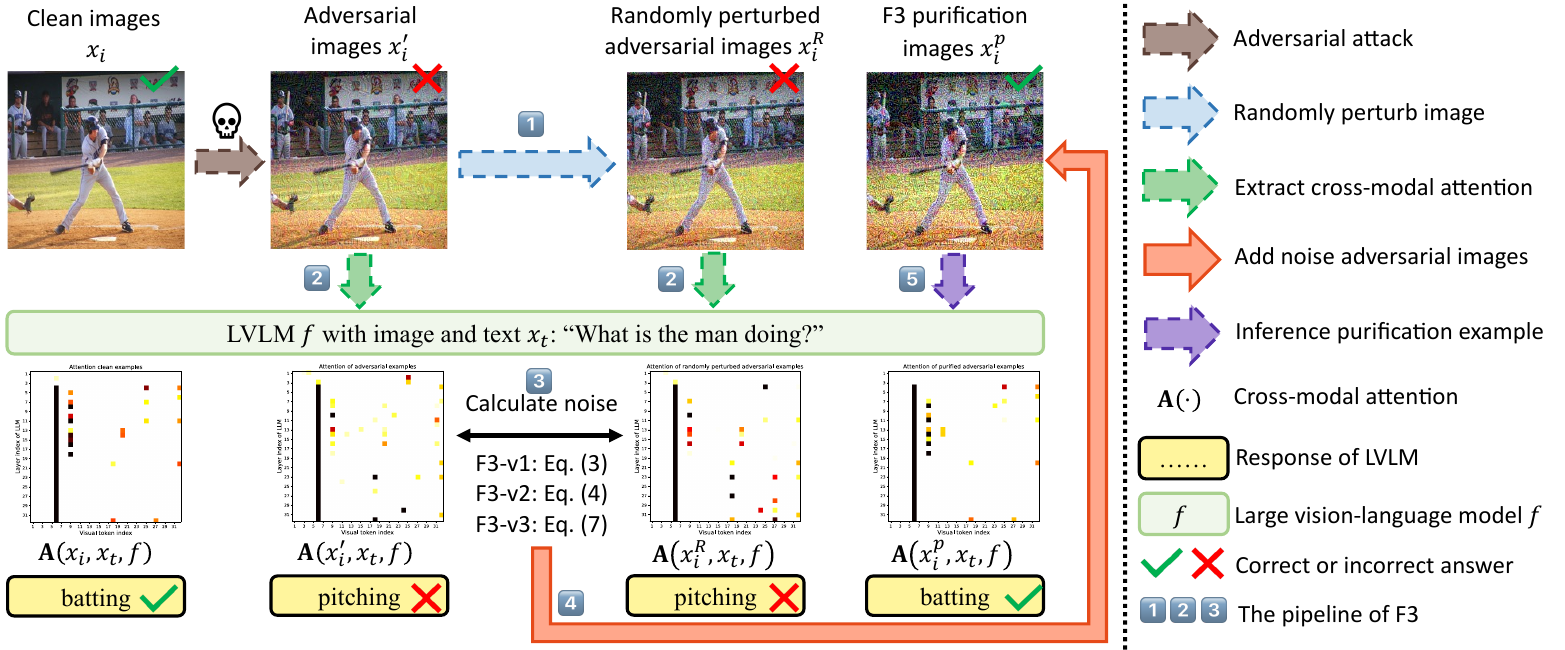}
  \caption{Our training-free and efficient F3 effectively targets the reference attention $\mathbf{A}(x_i^R, x_t, f)$, by incorporating additional simple perturbations into adversarial examples. Interestingly, the continued addition of noise to these adversarial examples in this context paradoxically enhances their performance, paralleling the strategy of fighting fire with fire.}
  \label{fig:pipeline}
\end{figure*}

As shown in \cref{sec:clean_attention}, \Cref{eq:clean_loss} can effectively purify adversarial examples with accessable clean attention. However, \Cref{eq:clean_loss} reveals a critical limitation: the attention of a clean example cannot be determined without direct access to the clean example itself. Given this challenge in identifying clean attention, we explored incorporating random noise into adversarial examples, as described in \cref{eq:random_loss}. Our experiments revealed an intriguing phenomenon: while the randomly perturbed adversarial example $x_i^R$ remained unable to produce correct answers after the addition of noise, its cross-modal attention became significantly more aligned with clean attention. For instance, as illustrated in \cref{fig:pipeline}, the 8th token of $\mathbf{A}(x_i^R)$ showed a substantial increase compared to the adversarial attention $\mathbf{A}(x_i')$, which is more similar to the clean attention $\mathbf{A}(x_i)$. This suggests that $\mathbf{A}(x_i^R)$ provides a closer approximation to the inaccessible clean attention $\mathbf{A}(x_i)$.
\begin{equation}
    \label{eq:random_loss}
    x_i^R = \mathcal{R}(x_i', \alpha_\infty)= x_i' - \alpha, \alpha\sim\mathcal{U}[-\alpha_\infty, \alpha_\infty].
\end{equation}

We provide quantitative analysis in \cref{tab:score_of_clean_adv}. By introducing noise bounded by $\alpha_\infty$ to adversarial examples, we evaluated both their performance and attention similarity before and after noise addition as \cref{eq:random_loss}. \textbf{While the addition of random noise did not significantly improve VQA scores, it notably aligned the attentions more closely with those of clean examples.} Furthermore, as the intensity of the added noise $\alpha_\infty$ increased, the attentions further approximated clean attention. Although random noise perturbation does not fully purify adversarial examples, the resulting cross-modal attention suggests a promising direction for purification. This approach serves as a practical approximation for ideal yet inaccessible clean attention, which we refer to as \emph{reference attention} ($\mathbf{A}(x_i^R)$) for the purpose of purification.

\begin{table}[!htbp]
  \caption{The VQA scores and attention similarity among clean, adversarial, and randomly perturbed adversarial examples under C\&W unadaptive attacks.}
  \label{tab:score_of_clean_adv}
  \centering
  \resizebox{\linewidth}{!}{
  \begin{tabular}{c|c|c|ccccc}
    \toprule
    \multirow{2}{*}{} & \multirow{2}{*}{Clean} & \multirow{2}{*}{Adv} & \multicolumn{5}{c}{F3-v1($\alpha_\infty$) as \cref{eq:random_loss}} \\
    \cmidrule(lr){4-8}
    & & & 2/255 & 4/255 & 8/255 & 16/255 & 32/255 \\
    \midrule
    VQA scores $\uparrow$ & 75.95 & 24.88 & 24.79 & 24.88 & 25.01 & 24.80 & 31.34 \\
    MSE($\mathbf{A}_\text{clean}, \mathbf{A}$) $\downarrow$ & 0 & 16.03 & 16.06 & 15.99 & 15.85 & 14.92 & 12.02 \\
    KL($\mathbf{A}_\text{clean}, \mathbf{A}$) $\downarrow$ & 0 & 4.91 & 4.91 & 4.89 & 4.83 & 4.46 & 3.34 \\
    \bottomrule
  \end{tabular}
  }
\end{table}

\subsection{F3-v2: Purifying Adversarial Examples towards the Reference Attention}
\label{sec:random_noise_ref_purify}

As demonstrated in \cref{sec:random_noise_attention}, adversarial examples with random perturbations exhibit attention patterns that are closer to those of clean examples compared to traditional adversarial examples. For a given adversarial example $x_i'$, by introducing random noise, we can derive $x_i^R=\mathcal{R}(x_i', \alpha_\infty)$ as detailed in \cref{eq:random_loss}, where $\alpha_\infty$ imposes a constraint on the intensity of the added noise. By optimizing the cross-modal attention of adversarial examples to align more closely with that of randomly perturbed examples, we can obtain a purified adversarial example, denoted as $x_i^p$, as described in \cref{eq:only_direction_loss}. In this equation, $\beta_\infty$ represents the perturbation limit.
\begin{align}
    \label{eq:only_direction_loss}
    x_i^p = x_i'-\beta\times\text{sign}(\nabla_{x_i'}||&\mathbf{A}(x_i', x_t, f)- \mathbf{A}(\mathcal{R}(x_i', \alpha_\infty),x_t, f)||_2), \\
    \beta&\sim\mathcal{U}[0, \beta_\infty]. \notag
\end{align}

Compared to F3-v1, F3-v2 introduces a constraint solely on the direction of the perturbation, specifically within the $\text{sign}(\cdot)$ function. In \cref{eq:only_direction_loss}, $\alpha_\infty$ represents the intensity of the random noise added to the adversarial example for estimating clean attention, while $\beta_\infty$ denotes the maximum perturbation intensity applied during the purification process based on the estimated direction of clean attention (\ie, reference attention). We explored various combinations of $\alpha_\infty$ and $\beta_\infty$ values to purify adversarial examples using F3-v2. In \cref{tab:score_of_random_guidence}, each row corresponds to a specific $\alpha_\infty$ value used to derive the clean attention estimate, while each column represents a specific $\beta_\infty$ value applied during the purification process. As shown in \cref{tab:score_of_random_guidence}, \textbf{purification performance can be significantly enhanced by simply constraining the direction (positive or negative) of the noise}, as opposed to employing a ``random direction'' strategy in F3-v1. Within a specific range, increasing $\alpha_\infty$ improves the precision of clean attention estimation, thus enhancing purification efficacy. This finding emphasizes the crucial role of controlling perturbation directions in achieving optimal results. Notably, setting $\alpha_\infty=16/255$ for clean attention estimation and maintaining a noise limit of $\beta_\infty=32/255$ during adversarial example purification yields the best outcomes.

\begin{table}[!hbtp]
  \caption{Purification results of F3-v2/v3($\alpha_\infty, \beta_\infty$) as \cref{eq:only_direction_loss,eq:x_fangda}. The $\alpha_\infty$ denotes the noise intensity used to obtain the reference attention, and $\beta_\infty$ denotes the noise intensity used for purification. In F3-v1, we directly use randomly perturbed adversarial examples as purification examples.}
  \label{tab:score_of_random_guidence}
  \centering
  \small
  % \resizebox{0.9\linewidth}{!}{
  \begin{tabular}{c|c|ccccc}
    \toprule
    & \multirow{2}{*}{$\alpha_\infty$} & \multicolumn{5}{c}{$\alpha_\infty$ in F3-v1 or $\beta_\infty$ in F3-v2/v3} \\
    \cmidrule(lr){3-7}
     & & 2/255 & 4/255 & 8/255 & 16/255 & 32/255 \\
    \midrule
    \multicolumn{2}{c|}{F3-v1} & 24.79 & 24.88 & 25.01 & 24.80 & 31.34 \\
    \midrule
    \multirow{5}{*}{F3-v2}& 2/255 & 26.20 & 27.48 & 30.19 & 35.57 & 44.59 \\
    & 4/255 & 26.40 & 27.19 & 30.36 & 36.41 & 45.50 \\
    & 8/255 & 27.01 & 28.14 & 31.42 & 35.88 & 45.15 \\
    & 16/255 & 27.86 & 28.56 & 31.91 & 37.16 & 46.15 \\
    & 32/255 & 27.68 & 28.31 & 31.00 & 35.97 & 45.32 \\
    \midrule
    \multirow{5}{*}{F3-v3} & 2/255 & 26.77 & 28.82 & 34.66 & 43.69 & 52.33 \\
    & 4/255 & 27.15 & 29.53 & 35.55 & 44.46 & 53.56 \\
    & 8/255 & 26.77 & 30.17 & 36.06 & 45.74 & 55.42 \\
    & 16/255 & 28.02 & 30.36 & 35.99 & 45.39 & 54.74 \\
    & 32/255 & 28.11 & 30.66 & 36.41 & 45.48 & 54.33 \\
    \bottomrule
  \end{tabular}
  % }
\end{table}

\subsection{F3-v3: Finer Control of Purifying Noise}
\label{sec:random_noise_ref_purify_fangda}

For F3-v2, we employ randomly perturbed adversarial examples exclusively to determine the direction of clean example attention, thereby guiding the optimization process (\ie, whether to increase or decrease) for each pixel during the purification of adversarial examples. However, this approach involves randomly selecting $\beta$ within predefined perturbation limits ($\beta_\infty$), which may not be optimal since uniformly applying random noise across all pixels is inherently unreasonable. When backpropagating the loss function to optimize toward the estimated clean attention for each pixel of the adversarial examples, the resulting gradient contains both directional and intensity information. Specifically, pixels with larger gradients should experience more significant perturbations, while those with smaller gradients should undergo subtler adjustments. The purification process is detailed in \cref{eq:grad,eq:grad_norm,eq:x_fangda}.
\begin{align}
    \label{eq:grad} g = \nabla_{x_i'}&||\mathbf{A}(x_i', x_t, f)-\mathbf{A}(\mathcal{R}(x_i', \alpha_\infty), x_t, f)||_2. \\
    \label{eq:grad_norm} g_\text{norm} &= \frac{(g-g_\text{min})}{(g_\text{max}-g_\text{min})}. \\
    \label{eq:x_fangda} x_i^p = x_i' - \beta_\infty\times&\max{\left(0, \min{\left(\frac{g_\text{norm}}{\text{avg}({g_\text{norm}})}, 1\right)}\right)}\times\text{sign}(g).
\end{align}
As shown in \cref{eq:grad}, the gradient $g$ is calculated with respect to reference attention using randomly perturbed adversarial examples. \Cref{eq:grad_norm} normalizes this gradient within the range $[0, 1]$. To address the significant variability in the normalized gradient values ($g_\text{norm}$), \cref{eq:x_fangda} further amplifies $g_\text{norm}$ by dividing it by its mean value $\text{avg}({g_\text{norm}})$, where $g_\text{max}$ and $g_\text{min}$ represent the maximum and minimum gradient values, respectively. Subsequently, noise is applied at each pixel location based on both the direction and magnitude of the gradient. 

The overall pipeline for our F3 approach is illustrated in \cref{fig:pipeline}. F3-v1, v2, and v3 produce noise and inject it into adversarial samples using \cref{eq:random_loss,eq:only_direction_loss,eq:x_fangda}, respectively. The results of adversarial purification using F3-v3 are presented in \cref{tab:score_of_random_guidence}. Compared to F3-v2, which only consider direction and randomly determine perturbation magnitude, the purification performance achieved by incorporating gradient information to control both the direction and intensity of perturbations is significantly improved. \textbf{This highlights the critical importance of our finer-grained control mechanism, which leverages both directional and intensity information in the F3 noise addition purification process}.

\section{In-depth Experiments and Analyses on F3}
\label{sec:experiment}

In \cref{sec:methods}, we have conducted preliminary experiments with F3, and we will make a comprehensive evaluation of F3 in \cref{sec:experiment} on different scenarios, including various LVLMs, attack methods, and datasets with multiple competitive purification methods. The detailed introductions of our settings are as follows.

\subsection{Experiment Setup}
\label{sec:experiment_setup}
\noindent
\textbf{Attack and defenses.} Please refer to Appendix.

\noindent
\textbf{LVLMs.} We selected several representative large-language vision models (LVLMs) for our experiments, including early-stage models such as BLIP-2 \cite{li2023blip} and InstructBLIP \cite{dai2024instructblip}, as well as iconic models like LLaMAv1.5 \cite{liu2024improved}. We placed particular emphasis on Qwen2.5-VL \cite{bai2025qwen2}, the current state-of-the-art (SOTA) open-source LVLM. While the primary experimental results in \cref{sec:methods} focus on InstructBLIP, we also conducted additional experiments across a broader range of LVLMs to validate the effectiveness and generalization capability of our proposed method, F3.

\noindent
\textbf{Datasets.} We primarily used the VQA v2 dataset \cite{goyal2017making} for evaluation. Additionally, we constructed a Q\&A dataset using ImageNet \cite{deng2009imagenet}. Beyond the Q\&A task, we also evaluated performance on the image captioning task using the COCO dataset \cite{lin2014microsoft}.

\subsection{Generalize F3 to Various LVLMs}
\label{sec:different_lvlm_f3}

\noindent
\textbf{Unadaptive attacks.} To validate whether F3 possesses a generalized adversarial purification capability across different LVLMs and attack methods, we conducted a comprehensive evaluation using several widely-adopted LVLMs. These models employ distinct architectures for vision-language projector, specifically Q-former and MLP-based structures. We assessed F3's robustness against two popular attack methods: the C\&W attack and AutoAttack. As detailed in \cref{tab:compare_diff_models_unadaptive}, our experimental results consistently demonstrate that F3 maintains strong purification performance across all tested LVLMs and attack methods, thereby confirming the generalizability and effectiveness of the F3 approach.

\begin{table}[!hbtp]
  \caption{The generalization capability of F3-v3 across various LVLMs under non-adaptive attack scenarios. Here, ``Adv'' represents the results obtained following an adversarial attack, while ``F3-v3'' represents the purified results, respectively.}
  \label{tab:compare_diff_models_unadaptive}
  \centering
  \small
  % \resizebox{\linewidth}{!}{
  \begin{tabular}{c|c|c|c|c}
    \toprule
    \multirow{2}{*}{Attack method} & \multirow{2}{*}{LVLM} & \multicolumn{3}{c}{VQA scores}\\
     & & Clean & Adv & F3-v3 \\
    \midrule
    \multirow{5}{*}{C\&W} & BLIP-2 XL & 56.49 & 16.40 & 42.81 \\
    & BLIP-2 XXL & 57.96 & 13.33 & 43.95\\
    \cmidrule(lr){2-5}
    & InstructBLIP XL & 73.11 & 19.12 & 53.69 \\
    & InstructBLIP XXL & 71.54 & 19.46 & 52.20 \\
    & InstructBLIP 13B & 61.59 & 20.86 & 49.43 \\
   \midrule
    \multirow{3}{*}{AutoAttack} & LLaVAv1.5 7B & 76.19 & 17.19 & 54.92 \\
    \multirow{3}{*}{($\epsilon_\infty=16$)}& LLaVAv1.5 13B & 77.37 & 17.54 & 55.55 \\
    \cmidrule(lr){2-5}
    & Qwen2.5-VL 3B & 80.47 & 22.66 & 47.58 \\
    \bottomrule
  \end{tabular}
  % }
\end{table}

\noindent
\textbf{Adaptive AutoAttack compared to classical purification methods}. It is essential for adversarial defense or purification methods to emphasize the importance of demonstrating their effectiveness under adaptive attack conditions \cite{athalye2018obfuscated, tramer2020adaptive}. As shown in \cref{tab:compare_diff_models_adaptive}, F3 demonstrates superior robustness against adaptive adversarial attacks compared to other methods such as SR \cite{mustafa2019image}, JPEG \cite{jia2019comdefend}, and R\&P \cite{xie2018mitigating}. This highlights its effectiveness in scenarios where attackers have full knowledge of the defense mechanisms. 

\begin{table}[!hbtp]
  \caption{The comparison of F3-v3 with different defense strategies under adaptive attacks. Notably, since Qwen2.5-VL employs dynamic resolution adjustment for input images, the corresponding adaptations for SR, JPEG, and R\&P under varying resolutions remain to be investigated.}
  \label{tab:compare_diff_models_adaptive}
  \centering
  \small
  % \resizebox{\linewidth}{!}{
  \begin{tabular}{c|c|c|c|c|c}
    \toprule
    \multirow{2}{*}{Model} & \multicolumn{5}{c}{VQA scores} \\
    & Clean & SR & JPEG & R\&P & F3-v3 \\
    \midrule
    LLaVAv1.5 7B & 76.19 & 26.56 & 33.02 & 43.08 & 60.23 \\
    LLaVAv1.5 13B & 77.37  & 27.81 & 37.81 & 41.52 & 62.60 \\
    \midrule
    Qwen2.5-VL 3B & 80.47 & - & - & - & 50.94 \\
    \bottomrule
  \end{tabular}
  % }
\end{table}

\begin{table}[!hbtp]
  \caption{Comparisons of VQA scores and inference time between F3-v3 and DiffPure on InstructBLIP-7B. Normalizing by the inference time of 7B LVLMs reveals that DiffPure's purification time increased nearly 50-fold, highlighting the efficiency of the F3-v3.}
  \label{tab:compare_diff_purify_method}
  \centering
  \small
  \begin{tabular}{c|c|c|c}
    \toprule
    Defense & \multicolumn{2}{c|}{Inference time (normalized)} & VQA scores \\
    method & NVIDIA A800 & NVIDIA H800 & robust \\
    \midrule
    No defense & 1.00 $\times$ & 1.00 $\times$ & 28.88  \\
    DiffPure & \textcolor{red!90}{48.3 $\times$} & \textcolor{red!90}{57.6 $\times$} & 61.64\\
    F3-v3 & \textcolor{ForestGreen}{\textbf{3.71 $\times$}} & \textcolor{ForestGreen}{\textbf{4.33 $\times$}} &  52.52\\
    \bottomrule
  \end{tabular}
\end{table}

\subsection{Compare F3 with DiffPure}

DiffPure \cite{nie2022DiffPure} employs a diffusion model-based approach, has established itself as the current leader in adversarial image purification. Although F3 demonstrates purification performance that is only slightly less effective than that of DiffPure, it is important to note that DiffPure presents several critical limitations that significantly impede its practical implementation in real-world purification scenarios. In contrast, our innovative F3 method, while still in the early stages of exploration, demonstrates significant potential as a more practical and effective solution: 

\noindent
\textbf{(1) Training-Free Design}. Unlike DiffPure, which relies on pre-trained diffusion models that require significant training costs and data resources, F3 operates in a completely training-free manner. This design allows F3 to be seamlessly integrated into LVLMs without additional training requirements, providing a more flexible and efficient solution. 

\noindent
\textbf{(2) Computational Efficiency}. The efficiency of a defense or purification method is critical for practical deployment. Our evaluation reveals that while F3 is slightly less robust than DiffPure, as shown in \Cref{tab:compare_diff_purify_method}, F3 introduces only 2-3 times the inference cost compared to an undefended baseline. This minimal overhead makes F3 a highly practical choice for real-world LVLMs, whereas DiffPure's higher computational demands (50 times the inference cost) render it less suitable for large-scale applications.

\noindent
\textbf{(3) Domain Agnosticism}. A key limitation of DiffPure is its dependence on pre-trained diffusion models that are tailored to specific data domains. For example, Score SDE \cite{song2021scorebased} is designed for CIFAR-10 \cite{krizhevsky2009learning}, Guided Diffusion \cite{dhariwal2021diffusion} for ImageNet \cite{deng2009imagenet}, and DDPM \cite{ho2020denoising} for CelebA-HQ \cite{karras2018progressive}. When the input data domain mismatches the pre-trained diffusion model, such as processing face images through a model trained on natural scenes, DiffPure often produces distorted outputs. This issue is particularly problematic for LVLMs, which must handle diverse task scenarios and data distributions, including chart understanding and document analysis. In contrast, F3 eliminates the need to select domain-specific purification models, providing a more versatile solution. Although our current evaluation on the VQA v2 dataset, which primarily consists of natural images within DiffPure's original distribution, does not fully expose this limitation, it remains a critical concern for broader applications.

\noindent
\textbf{(4) Suitable for dynamic resolution LVLMs}. DiffPure is built upon pre-trained diffusion models and is specifically designed for LVLMs with fixed resolution outputs. However, state-of-the-art LVLMs such as Qwen2.5-VL \cite{bai2025qwen2} predominantly employ dynamic resolution approaches, which limits the applicability of DiffPure in these advanced models. In contrast, F3 eliminates dependence on input resolution, thereby offering greater deployment flexibility across various scenarios.

\subsection{Generalize F3 to Various Datasets and Tasks}
\label{sec:different_dataset_f3}

\noindent
\textbf{Evaluating F3 on ImageNet}. To comprehensively assess the generalization capability of our F3 framework, we constructed a Q\&A dataset by selecting images from ImageNet-1K \cite{deng2009imagenet} and generating corresponding questions. As shown in \cref{tab:compare_diffpure}, when applied to unadaptive C\&W attacks on ImageNet-1K, F3 demonstrates strong purification performance while minimizing harm to clean images.

\begin{table}[!hbtp]
  \caption{The robust VQA scores of F3-v3 on ImageNet.}
  \label{tab:compare_diffpure}
  \centering
  \small
  \resizebox{0.92\linewidth}{!}{
  \begin{tabular}{c|c|c|cc}
    \toprule
    Attack & No-defense & \multirow{2}{*}{Diffpure \cite{nie2022DiffPure}} & \multicolumn{2}{c}{F3-v3 ($\alpha_\infty=16, \beta_\infty$)} \\
    \cmidrule(lr){4-5}
    method & (w/o purify) & &24/255 & 32/255 \\
    \midrule
    Clean & 81.5\% & 62.3\% & \textbf{74.7\%} & \textbf{72.8\%} \\ 
    C\&W & 23.8\% & 59.9\% & \textbf{62.8\%} & \textbf{65.1\%} \\
    \bottomrule
  \end{tabular}
  }
\end{table}

\noindent
\textbf{Scaling F3 to Larger VQA Datasets}. To further validate the robustness of F3, we evaluated F3 on an expanded dataset $\mathcal{D}_\text{VQAv2}^{5000}$ instead of $\mathcal{D}_\text{VQAv2}^{1000}$. \Cref{tab:score_of_vqa_5000} indicates that F3 maintains consistent performance even when scaled to larger datasets.

\begin{table}[!htbp]
  \caption{VQA scores on $\mathcal{D}_\text{VQAv2}^{5000}$ under non-adaptive attacks.}
  \label{tab:score_of_vqa_5000}
  \centering
  \resizebox{0.88\linewidth}{!}{
  \begin{tabular}{c|c|ccccc}
    \toprule
    \multirow{2}{*}{Clean} & \multirow{2}{*}{Adv} & \multicolumn{5}{c}{F3-v3 ($\alpha_\infty=16, \beta_\infty$)}\\
    \cmidrule(lr){3-7}
    & & 2/255 & 4/255 & 8/255 & 16/255 & 32/255 \\
    \midrule
    75.93 & 24.19 & 28.24 & 31.33 & 36.64 & 44.94 & 54.71 \\
    \bottomrule
  \end{tabular}
  }
\end{table}

\noindent
\textbf{Exploring F3 in image captioning tasks}. While our primary focus was on the Q\&A task, we expanded our investigation to evaluate F3's performance on image captioning tasks using the COCO dataset \cite{lin2014microsoft}. In this study, we still concentrated on the cross-modal attention for the first generated token, achieving promising outcomes as presented in \cref{tab:caption_tasks}. Building on these encouraging results, further refinement of F3 to address each generated token individually could potentially yield even greater improvements. This initial exploration underscores F3's broader applicability and highlights its potential utility across additional tasks.

\begin{table}[!htbp]
  \caption{Evaluate F3-v3 under adaptive attacks on image captioning task and COCO dataset.}
  \label{tab:caption_tasks}
  \centering
  \resizebox{\linewidth}{!}{
  \begin{tabular}{c|ccccc}
    \toprule
    Method & CIDEr & BLEU-1 & ROUGE-L & METEOR & SPICE \\
    \midrule
    Clean & 154.5 & 83.6 & 61.6 & 31.8 & 25.4\\
    No-defense & 99.5 & 66.7 & 47.8 & 25.1 & 19.0 \\
    R\&P & 105.3 & 69.0 & 50.2 & 25.2 & 19.3\\
    F3-v3 & \textbf{116.5} & \textbf{73.3} & \textbf{53.8} & \textbf{25.9} & \textbf{20.2} \\
    \bottomrule
  \end{tabular}
  }
\end{table}

\subsection{Are the Cross-modal Attention of Purified Examples Cleaner than Adversarial Ones?}
\label{sec:attention_compare}
In \cref{sec:random_noise_attention,sec:random_noise_ref_purify,sec:random_noise_ref_purify_fangda}, we introduced three distinct methods: F3-v1, F3-v2, and F3-v3. The results presented in \cref{tab:score_of_random_guidence} demonstrate a clear hierarchy in purification performance, with F3-v3 surpassing F3-v2, which in turn outperforms F3-v1 (F3-v3 > F3-v2 > F3-v1). Consistent with our theoretical motivation, these methods offer progressively finer control over the purifying noise. This refinement leads to enhanced alignment between the cross-modal attention of purified examples and that of clean examples. Specifically, the similarity in attention patterns follows the same hierarchy: F3-v3 > F3-v2 > F3-v1. \Cref{tab:similarity_of_attneion_of_three_sections} quantitatively confirms this relationship, providing empirical evidence that supports our theoretical framework. Furthermore, as $\beta_\infty$ increases, the purification attention matrix $\mathbf{A}$ becomes increasingly indistinguishable from the clean attention matrix $\mathbf{A}_\text{clean}$, further validating our analysis.

\begin{table}[!htbp]
  \caption{The attention similarity of F3-v1, F3-v2, F3-v3. As we analysed, F3-v3 performed best, with its attention closest to clean attention than F3-v2 and F3-v1.}
  \label{tab:similarity_of_attneion_of_three_sections}
  \centering
  \resizebox{\linewidth}{!}{
  \begin{tabular}{c|c|c|c|c}
    \toprule
    & $\beta_\infty$ & VQA scores & MSE($\mathbf{A}_\text{clean}, \mathbf{A}$) & KL($\mathbf{A}_\text{clean}, \mathbf{A}$) \\
    \midrule
    \multicolumn{2}{c|}{Clean image} & 75.95 & 0 & 0 \\
    \multicolumn{2}{c|}{Adversarial image} & 24.88 & 16.03 & 4.91 \\
    \midrule
    \multicolumn{2}{c|}{F3-v1}& 24.80 & 14.92 & 4.46 \\
    \midrule
    \multirow{3}{*}{F3-v2} & 8/255 & 31.91 & 12.16 & 3.72 \\
    & 16/255 & 37.16 & 11.04 & 3.45 \\
    & 32/255 & 46.15 & 9.16 & 2.87 \\
    \midrule
    \multirow{3}{*}{F3-v3} & 8/255 & 35.99 & 11.35 & 3.53 \\
    & 16/255 & 45.39 & 9.69 & 3.03 \\
    & 32/255 & 54.74 & 7.82 & 2.53 \\
    \bottomrule
  \end{tabular}
  }
\end{table}

\subsection{Measuring Possible Negative Impact of F3 on Clean Examples}
\label{sec:clean_drop} 

There is typically a trade-off between an LVLM's performance on clean examples and its robustness against adversarial examples. In the context of adversarial purification, this balance shifts to minimizing negative impacts on clean examples while enhancing purification performance for adversarial ones. As shown in \cref{tab:compare_impact_clean}, various F3 settings influence both clean and adversarial outcomes. While stronger F3 configurations improve adversarial purification performance, they also tend to degrade results on clean examples. Specifically, applying F3 resulted in a 10-point decrease in clean VQA scores but delivered a significant 30-point improvement in adversarial VQA scores.

\begin{table}[!hbtp]
  \caption{The negative impact of F3-v3 on clean examples.}
  \label{tab:compare_impact_clean}
  \centering
  \small
  \begin{tabular}{l|c|ccc}
    \toprule
    \multirow{2}{*}{Dataset} & w/o purify & \multicolumn{3}{c}{F3-v3($\alpha_\infty=16, \beta_\infty$)}\\
    \cmidrule(lr){3-5}
    & by F3 & 8/255 & 16/255 & 32/255 \\
    \midrule
    Clean images & 75.95 & 68.10 & 66.27 & 63.26 \\
    C\&W images & 24.88 & 35.99 & 45.39 & 54.74 \\
    \bottomrule
  \end{tabular}
\end{table}

\subsection{Utilizing Multi-step Iterations in Adding-Perturbation Purification}
\label{sec:f3_multi_step}

In \cref{sec:methods}, the F3-v3 framework employs a single-step perturbation strategy for purifying adversarial examples. A natural question arises: can purification performance be enhanced by implementing multiple smaller iterative steps instead of a single larger step? To investigate this, we extend the F3-v3 process through multi-step iterations, and present the results in \cref{tab:score_of_multistep}, where $\epsilon_\infty$ denotes the total perturbation budget and $K$ represents the number of iteration steps. Through this process, multi-step iterations may exhibit backtracking in certain dimensions, which can reduce the perturbation amount. However, as the perturbation quantity is crucial for effective purification, we must carefully analyze these dynamics. To ensure fair comparisons, we measure the perturbation quantity using the $l_1$-norm and compare settings with similar $l_1$ norms. As shown in \cref{tab:score_of_multistep}, the multi-step strategy achieves superior performance compared to the single-step approach even when maintaining comparable $l_1$-norm perturbation levels. This suggests that the multi-step strategy enables more efficient purification by allowing finer-grained control over the direction of perturbations within the same total perturbation budget. Although we have not yet conducted an in-depth investigation of multi-step F3-v3 strategies, our preliminary studies have already demonstrated the promising capabilities of the F3 within the domain of adversarial purification. 

\vspace{-1em}

\begin{table}[!htbp]
  \caption{The VQA scores of multi-step F3-v3 purification, where $\epsilon_\infty$ denotes the total perturbation budget and $K$ represents the number of iteration steps.}
  \label{tab:score_of_multistep}
  \centering
  \small
  % \resizebox{\linewidth}{!}{
  \begin{tabular}{c|ccc|c|c}
    \toprule
    $K$ & $\beta_\infty$ & $\alpha_\infty$ & $\epsilon_\infty$ & VQA scores & $l_1$-norm\\
    \midrule
    1 & 6/255 & 16/255 & 16/255 & 34.61 & 5.75\\
    4 & 4/255 & 16/255 & 16/255 & 40.67 & 5.84 \\
    \midrule
    1 & 8/255 & 16/255 & 16/255 & 35.99 & 7.64 \\
    8 & 4/255 & 16/255 & 16/255 & 45.74 & 7.60 \\
    \bottomrule
  \end{tabular}
  % }
\end{table}

\section{Conclusion}
\label{sec:conclusion}

In this study, we investigate the fundamental relationship between cross-modal attention mechanisms and adversarial examples in LVLMs. We present F3, a novel framework that estimates clean attention direction by leveraging randomly perturbed adversarial examples. Our method achieves robustness by optimizing adversarial attention to better align with reference attention through Deliberate introduction of perturbations for adversarial purification. Extensive experiments demonstrate the effectiveness of our approach across multiple popular LVLMs (BLIP-2, InstructBLIP, LLaVAv1.5, Qwen2.5-VL) and diverse attack methods (C\&W, AutoAttack). Despite requiring significantly less computational overhead, F3 achieves comparable robustness evaluation metrics to DiffPure, which is resource-intensive with time-consuming diffusion processes. By addressing this critical yet previously underexplored dimension of adversarial purification in LVLMs, our training-free and computationally efficient framework not only enhances model robustness and security but also establishes new research directions for developing more resilient LVLM architectures. 

\clearpage
\newpage

\begin{acks}
This work was supported by Young Elite Scientists Sponsorship Program by CAST (2023QNRC001), the National Key R\&D Program of China (2023YFB4502200), the National Natural Science Foundation of China (No. 62376024, 62325405, 62104128, 62203257, 62031017, 62406159, U21B2031), Tsinghua University Initiative Scientific Research Program,  Beijing National Research Center for Information Science, Technology (No. BNR2024TD03001) and Beijing Innovation Center for Future Chips.
\end{acks}


\bibliographystyle{ACM-Reference-Format}
\balance
\bibliography{sample-base}

\clearpage
\newpage
\appendix

\section{Experimental setup for attack and defense}
\label{sec:setup_attack_defense}

\noindent
\textbf{Attacks.} For evaluating the visual adversarial purification method F3, we focused on adversarial attack scenarios targeting visual inputs within LVLMs. Our analysis specifically concentrated on attacks against visual modalities, excluding those targeting textual inputs \cite{lu2023set, zhang2022towards} or non-adversarial objectives such as privacy concerns and jailbreaking attempts \cite{liu2024safety, liu2024survey, fan2024unbridled}. To ensure a comprehensive evaluation, we selected two widely used adversarial attack methods in the field: C\&W and AutoAttack. For the C\&W attack \cite{carlini2017towards}, no specific constraints were imposed during implementation. Regarding AutoAttack \cite{croce2020reliable}, we employed two distinct versions to assess different attack strategies. The adaptive version utilized the ``rand'' configuration with $\epsilon_\infty=8$, designed to target random defense mechanisms over 20 attack steps and 10 Expectation Over Transformation (EOT) steps. Conversely, the non-adaptive version adopted the ``standard'' configuration with $\epsilon_\infty=16$.

\noindent
\textbf{Defenses.} To evaluate F3 against existing defense strategies in the context of visual adversarial attacks on LVLMs, we implemented three primary approaches: random resizing and padding (R\&P), super-resolution reconstruction (SR), and JPEG compression (JPEG). Additionally, we included DiffPure, a state-of-the-art method for adversarial image purification based on diffusion models. Despite its effectiveness, DiffPure presents significant practical challenges for deployment in real-world LVLM inference pipelines due to its high computational overhead and fixed output resolution constraints inherent to diffusion models. These limitations make it particularly challenging to integrate DiffPure with modern LVLMs, which increasingly process high-resolution inputs and already incur substantial computational costs. Our comparative analysis of F3 against these defense strategies primarily focuses on adaptive attack scenarios, but of course includes non-adaptive attacks.

\section{Limitation}

We summarize the limitations of our paper as follows:

(1) The design of the purification noise introduced by F3 is relatively straightforward, consisting of a single-step noise that is not meticulously calibrated in terms of size and direction. Despite this simplicity, F3 demonstrates commendable performance in countering the effects of noise. This indicates that even basic noise designs can be effective in certain contexts, highlighting the robustness of the F3 approach. In the future, we will further explore finer control of the F3 purification noise to further enhance F3.

(2) The effectiveness of the adaptive attack can be further optimized through various strategies, such as employing a combination of multiple F3 noises or integrating F3 with other defense or decontamination methods. Given the efficiency of our approach, these avenues present promising opportunities for enhancing F3's performance. Exploring these directions could lead to significant improvements in robustness and effectiveness against adversarial attacks.

(3) F3 focuses on optimizing the cross-modal attention of the first generated token. However, extending this mechanism to all subsequent tokens could potentially enhance its performance, particularly in captioning tasks. This represents a promising direction for future research. As demonstrated by the results in \cref{tab:caption_tasks}, even limited to the cross-modal attention of the first token, F3 successfully improves the robustness of the captioning tasks.

\section{More Details, Results and Analysis for F3}

\subsection{Detailed Experimental Setup for Adaptive Attack Evaluation}

Given the randomized nature of defense methods, we employ the rand version of AutoAttack for evaluation, including APGD-ce and APGD-dlr. The default configuration consists of 100 steps with EOT=20, resulting in up to $2\times 100 \times20=4,000$ forward and backward passes per sample. As shown in \cref{tab:compare_diff_purify_method}, DiffPure's forward time is approximately 50 times longer than normal. Consequently, executing adaptive AutoAttack on InstructBLIP Vicuna-7B with DiffPure requires substantial computation: 25 hours on an H800 GPU and 38 hours on an NVIDIA A800 GPU. To maintain practical experiment durations, we utilized a modified version of AutoAttack with 20 attack steps and EOT=10. 

\subsection{Distribution of question types in subdatasets}

We present in \cref{tab:question_type_num} the distribution of question types in the subdataset obtained through our sampling process. These subdatasets were sourced through direct sampling from the VQA v2 dataset.

\begin{table}[!htbp]
  \caption{The distribution of question types in the sampling dataset.}
  \label{tab:question_type_num}
  \centering
  \begin{tabular}{c|c|ccc}
    \toprule
    \multirow{2}{*}{Dataset} & \multirow{2}{*}{Total number} & \multicolumn{3}{c}{VQA v2 question type} \\
    \cmidrule(lr){3-5}
    & & yes/no & number & other \\
    \midrule
    $\mathcal{D}_\text{VQAv2}^{1000}$ & 1000 & 38.4\% & 14.0\% & 47.6\% \\
    \midrule
    $\mathcal{D}_\text{VQAv2}^{5000}$ & 5000 & 36.7\% & 13.0\% & 50.3\% \\
    \bottomrule
  \end{tabular}
\end{table}

\subsection{Are the Attacks We Use Strong Enough?}
\label{sec:attack_strong}
In our previous experiments, we employed three attack methods: PGD, C\&W, and AutoAttack (both ``rand'' and ''standard'' versions). Generally, adversarial attacks achieve an attack success rate (ASR) of 80\% or higher. However, in our work, the drop in VQA scores is not as significant. To clarify any potential misunderstandings regarding the strength of our attack methods, it is important to distinguish between our evaluation based on VQA scores and our evaluation based on ASR.

In previous adversarial attacks targeting classification tasks, performance was typically evaluated using ASR, where an attack is considered successful if the output category differs from the original. However, this approach does not directly translate to more complex multimodal VQA tasks. For instance, consider a question about a streetlight pole with a sign in an image: the VQA scoring metric \footnote{https://visualqa.org/evaluation.html}  considers multiple answers correct (\eg, ``light'', ``streetlight'', ``sign'', and ``light pole''), while ASR treats any inconsistency between pre-attack and post-attack answers as a successful attack. As a result, VQA scores provide a more nuanced measurement, and an apparent success in terms of ASR does not necessarily correspond to a significant decrease in VQA performance.

To comprehensively evaluate the attacks, we measured both the VQA scores and the ASR before and after applying the attacks. The results presented in \cref{tab:compare_with_vqa_asr} demonstrate that the attacks are sufficiently strong and that our purification approach is effective.

\begin{table}[!hbtp]
  \caption{Comparison of VQA scores and ASR. This proves that the attacks we used are effective enough.}
  \label{tab:compare_with_vqa_asr}
  \centering
  \small
  \begin{tabular}{c|c|c|c}
    \toprule
     \multirow{2}{*}{Attack method} &  \multicolumn{2}{c|}{VQA score} & \multirow{2}{*}{Attack success rate (ASR)} \\
     \cmidrule(lr){2-3}
      & Clean & Adv (Diff) \\
     \midrule
     PGD & 75.95 & 27.17 \textcolor{red}{\scriptsize (-48.78)} & 86.50\% \\
     C\&W & 75.95 & 24.88 \textcolor{red}{\scriptsize (-51.07)} & 90.00\% \\
     AutoAttack & 75.95 & 15.66 \textcolor{red}{\scriptsize (-60.29)} & 99.50\% \\
    \bottomrule
  \end{tabular}
\end{table}

We focus our investigation on visual adversarial examples within LVLMs, specifically targeting the visual modality rather than pursuing attacks that operate across both images and text simultaneously. This allows us to avoid employing multimodal attack frameworks such as Co-attack \cite{zhang2022towards}. Since we adopt a white-box attack approach, we also eliminate the need for methods like SGA \cite{lu2023set}, which are designed to enhance transferability across different models. Nonetheless, our selected adversarial attacks were sufficiently strong to validate the efficacy of F3 purification.

\subsection{Generalizability of F3 over Different Attack Methods and Configurations}
\label{sec:different_attack_method}

We also investigated other widely adopted attack methods beyond C\&W, including PGD (Projected Gradient Descent) \cite{madry2017towards} and AutoAttack \cite{croce2020reliable}. Specifically, for the PGD implementation, we conducted 20 iterations with a step size of $\epsilon=2/255$ and a maximum perturbation bound of $\epsilon=8/255$. For AutoAttack, we employed the default ``standard'' configuration ($\epsilon_\infty=8/255$). As shown in \cref{tab:compare_with_pgd}, our F3 method demonstrates robust performance in adversarial purification under both PGD and AutoAttack attacks.

\begin{table}[!hbtp]
  \caption{The generalizability of F3 to various adversarial attack methods. We fix $\alpha_\infty=16/255$.}
  \label{tab:compare_with_pgd}
  \centering
  \small
  % \resizebox{\linewidth}{!}{
  \begin{tabular}{c|c|c|ccc}
    \toprule
    \multirow{2}{*}{Attack method} & \multirow{2}{*}{Clean} & \multirow{2}{*}{Adv} & \multicolumn{3}{c}{F3-v3($\alpha_\infty=16, \beta_\infty$)}\\
    \cmidrule(lr){4-6}
    & & & 8/255 & 16/255 & 32/255 \\
    \midrule
    PGD \cite{madry2017towards} & 75.95 & 27.17 & 52.81 & 60.88 & 64.28 \\
    AutoAttack \cite{croce2020reliable} & 75.95 & 15.66 & 45.07 & 57.83 & 60.80 \\
    \bottomrule
  \end{tabular}
  % }
\end{table}

\subsection{Using Different Functions to Calculate the Noise via Cross-Modal Attention}
\label{sec:f3_kl_loss}
In \cref{tab:score_of_clean_adv}, we employed both MSE and KL metrics to measure the distance between clean attention and the reference attention. While we primarily used MSE in \cref{eq:only_direction_loss,eq:x_fangda}, we also explored KL divergence as an alternative. Specifically, since MSE treats all prediction errors equally and is equally sensitive to large and small errors, we considered KL divergence given that attention distributions inherently resemble probability distributions. 

For the attention tensor $\mathbf{A}_{L\times H\times M}$, we performed normalization over the visual token dimension ($M$) to ensure it conforms to the properties of a probability distribution at each layer ($L$) and for each multi-head attention ($H$). The results of using KL divergence instead of MSE in \cref{eq:x_fangda} for F3-v3 are presented in \cref{tab:compare_mse_kl}. Our experiments demonstrate that both MSE and KL losses yield effective and competitive performance in terms of purification accuracy.

\begin{table}[!htbp]
  \caption{Comparison of purification results using MSE and KL in F3-v3 and \cref{eq:x_fangda}.}
  \label{tab:compare_mse_kl}
  \centering
  \small
  \resizebox{\linewidth}{!}{
  \begin{tabular}{c|c|ccccc}
    \toprule
    \multirow{2}{*}{$\alpha_\infty$} & \multirow{2}{*}{Function in \cref{eq:x_fangda}} &  \multicolumn{5}{c}{$\beta_\infty$} \\
    \cmidrule(lr){3-7}
    & & 2/255 & 4/255 & 8/255 & 16/255 & 32/255 \\
    \midrule
    \multirow{2}{*}{8/255} & MSE & 26.77 & 30.17 & 36.06 & 45.74 & 55.42 \\
    & KL & 26.13 & 29.22 & 35.75 & 44.11 & 56.29 \\
    \midrule
    \multirow{2}{*}{16/255} & MSE & 28.02 & 30.36 & 35.99 & 45.39 & 54.74 \\
     & KL & 26.71 & 30.37 & 36.36 & 44.88 & 55.00 \\
    \bottomrule
  \end{tabular}
  }
\end{table}

\section{Additional Visualization Results of Images and Cross-Modal Attentions}
\label{sec:more_visual_result}

\begin{figure*}[htbp]
    \centering
    \begin{subfigure}{0.22\linewidth}
        \includegraphics[width=\linewidth]{figure/clean_image_000000327601.png}
        \caption{Clean image $x_i$.}
    \end{subfigure}
    \hfill
    \begin{subfigure}{0.22\linewidth}
        \includegraphics[width=\linewidth]{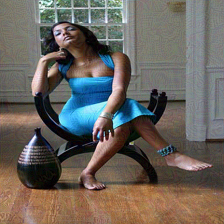}
        \caption{Adversarial image $x_i'$.}
    \end{subfigure}
    \hfill
    \begin{subfigure}{0.22\linewidth}
        \includegraphics[width=\linewidth]{figure/random_image_000000327601.png}
        \caption{Perturbed image $x_i^R$.}
    \end{subfigure}
    \hfill
    \begin{subfigure}{0.22\linewidth}
        \includegraphics[width=\linewidth]{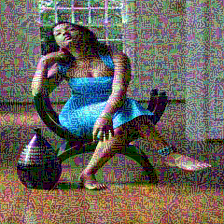}
        \caption{Purified image $x_i^p$.}
    \end{subfigure}
    \begin{subfigure}{0.22\linewidth}
        \includegraphics[width=\linewidth]{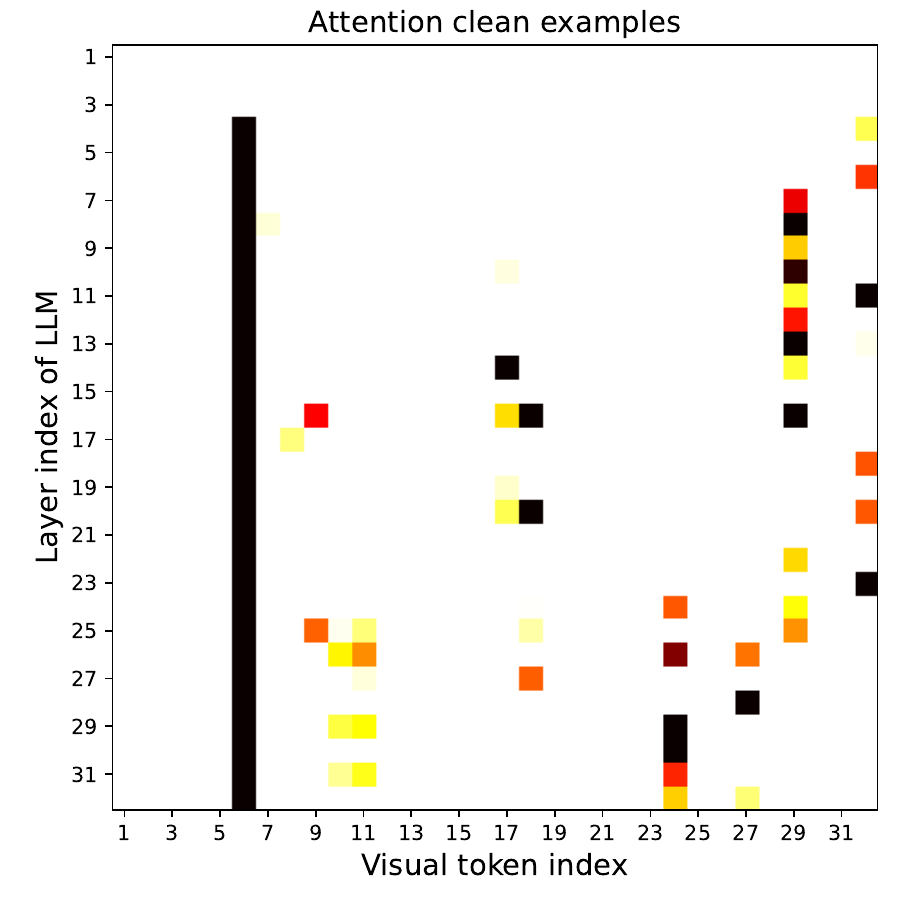}
        \caption{Clean attention $\mathbf{A}(x_i)$.}
    \end{subfigure}
    \hfill
    \begin{subfigure}{0.22\linewidth}
        \includegraphics[width=\linewidth]{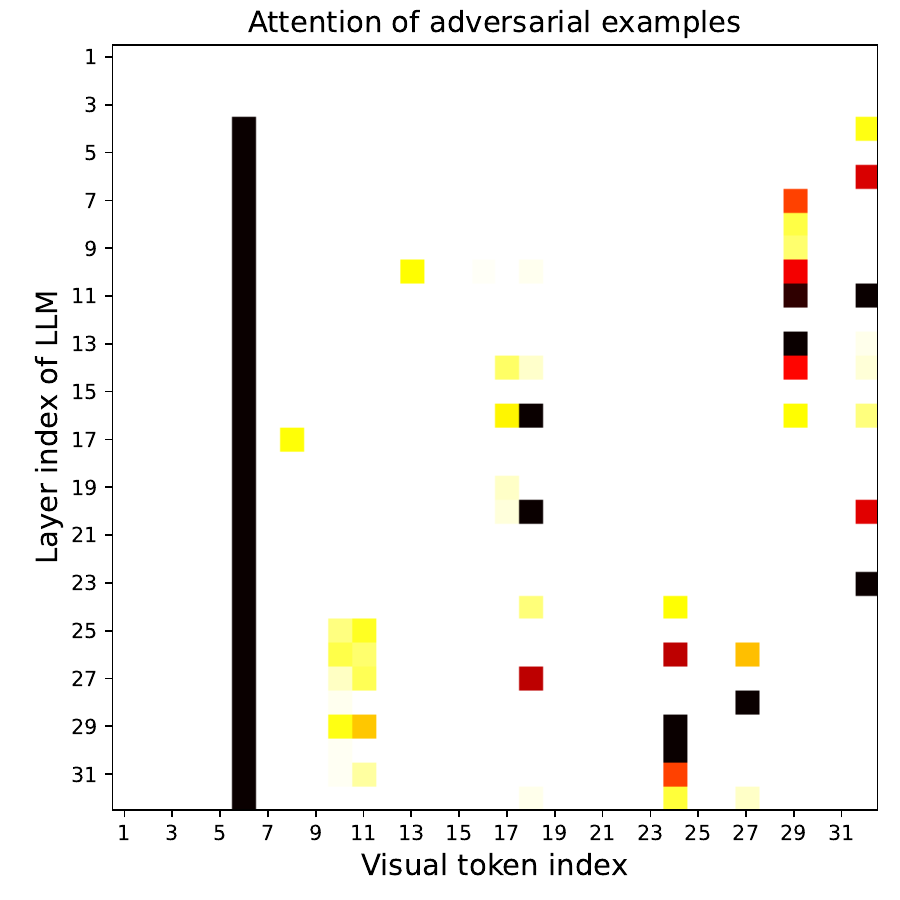}
        \caption{Adversarial attention $\mathbf{A}(x_i')$.}
    \end{subfigure}
    \hfill
    \begin{subfigure}{0.22\linewidth}
        \includegraphics[width=\linewidth]{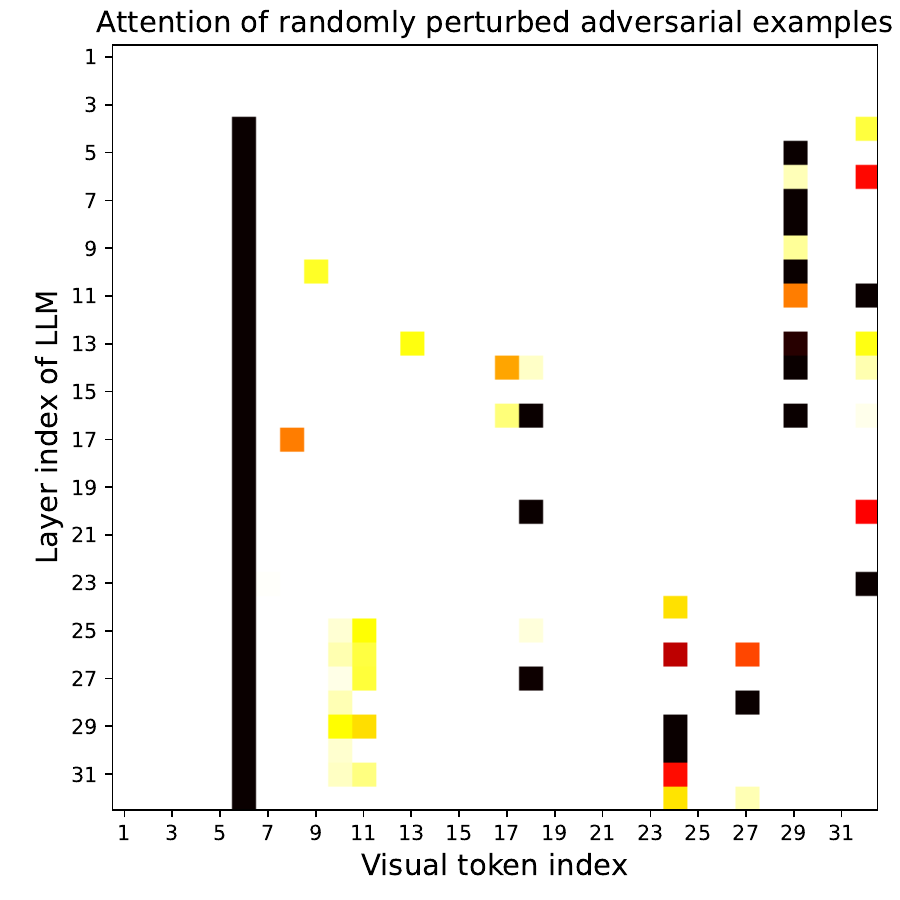}
        \caption{Reference attention $\mathbf{A}(x_i^R)$.}
    \end{subfigure}
    \hfill
    \begin{subfigure}{0.22\linewidth}
        \includegraphics[width=\linewidth]{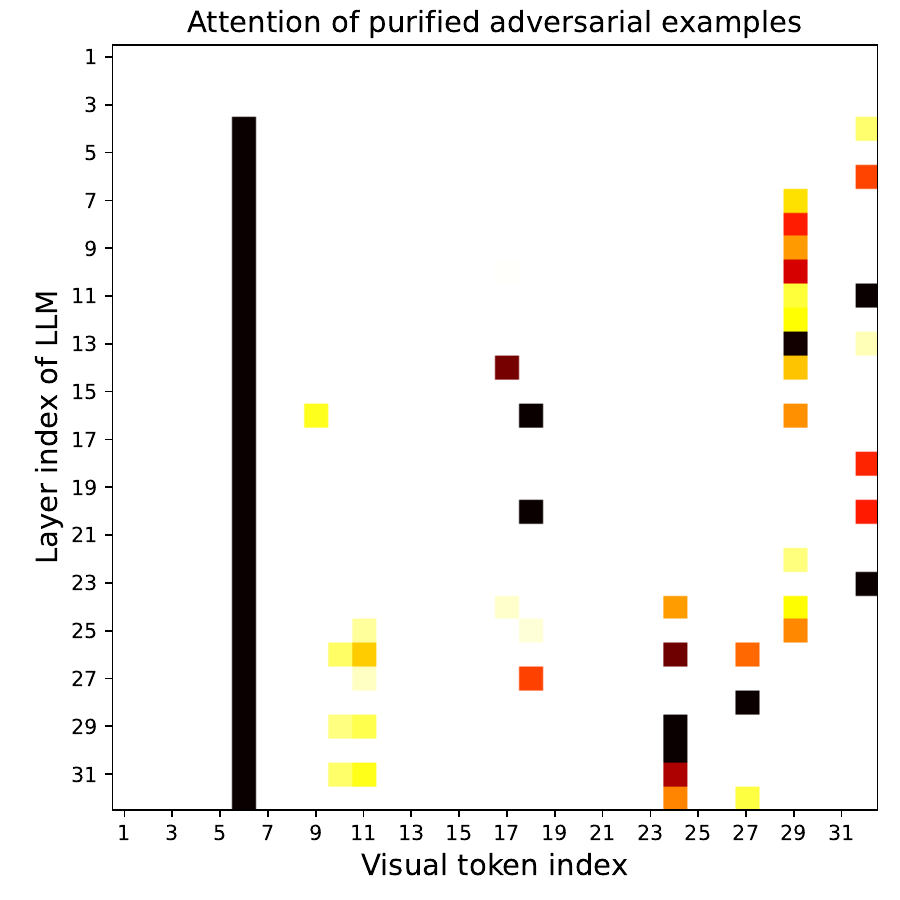}
        \caption{Purified attention $\mathbf{A}(x_i^p)$.}
    \end{subfigure}
    \caption{The visualization results under PGD attack. The question is ``What color is the women dress?''. The answers to the four images from (a) to (d) are ``blue'', ``green'', ``green'', and ``blue''.}
    \label{fig:attention_under_pgd}
\end{figure*}

\begin{figure*}[htbp]
    \centering
    \begin{subfigure}{0.22\linewidth}
        \includegraphics[width=\linewidth]{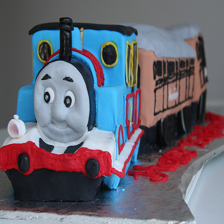}
        \caption{Clean image $x_i$.}
    \end{subfigure}
    \hfill
    \begin{subfigure}{0.22\linewidth}
        \includegraphics[width=\linewidth]{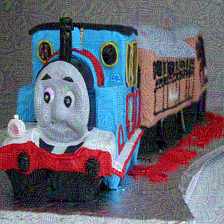}
        \caption{Adversarial image $x_i'$.}
    \end{subfigure}
    \hfill
    \begin{subfigure}{0.22\linewidth}
        \includegraphics[width=\linewidth]{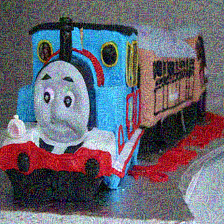}
        \caption{Perturbed image $x_i^R$.}
    \end{subfigure}
    \hfill
    \begin{subfigure}{0.22\linewidth}
        \includegraphics[width=\linewidth]{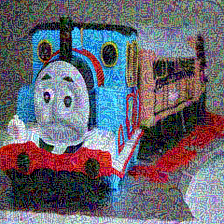}
        \caption{Purified image $x_i^p$.}
    \end{subfigure}
    \begin{subfigure}{0.22\linewidth}
        \includegraphics[width=\linewidth]{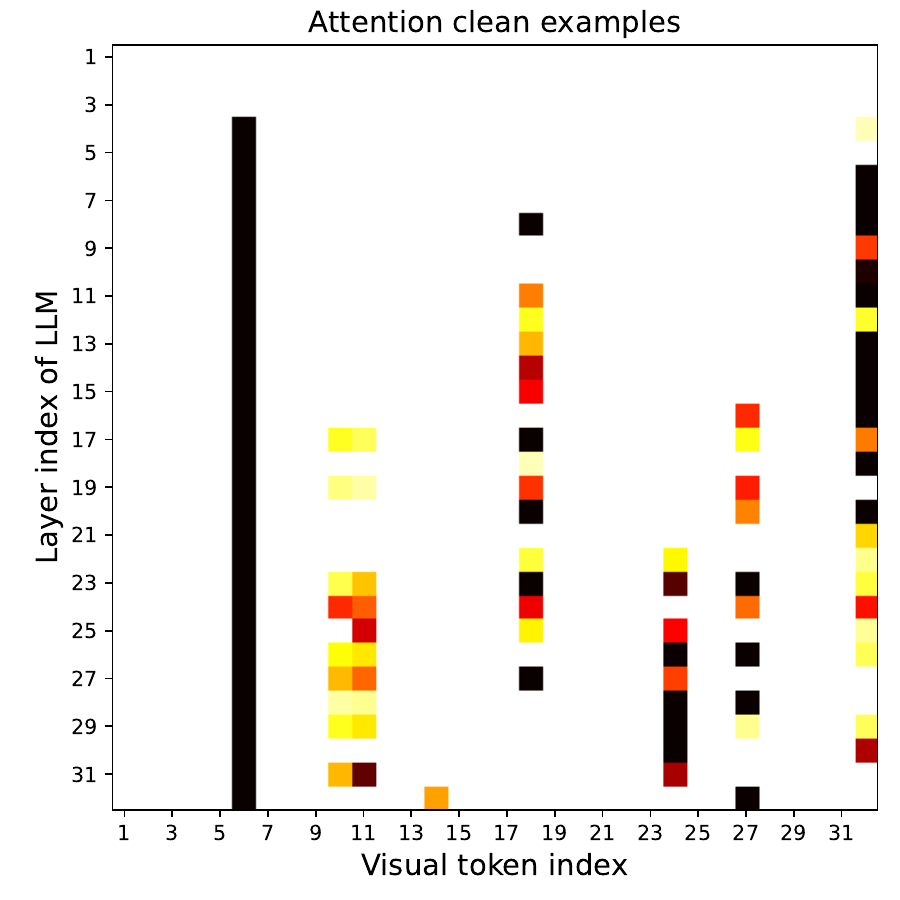}
        \caption{Clean attention $\mathbf{A}(x_i)$.}
    \end{subfigure}
    \hfill
    \begin{subfigure}{0.22\linewidth}
        \includegraphics[width=\linewidth]{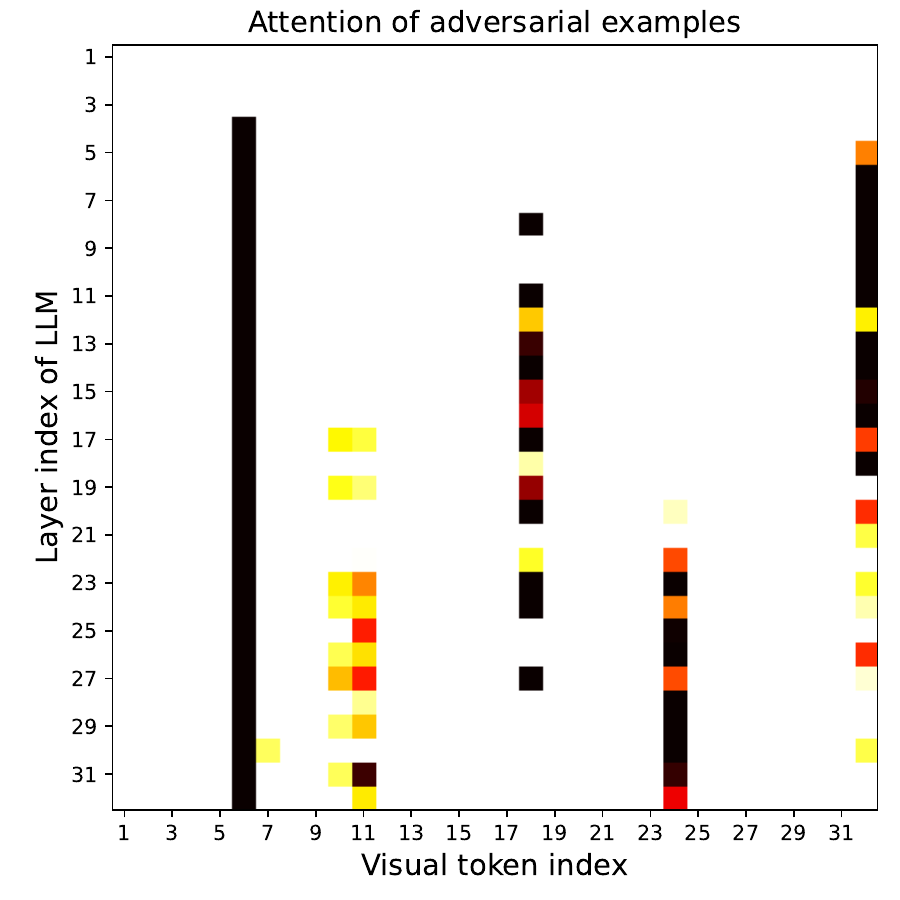}
        \caption{Adversarial attention $\mathbf{A}(x_i')$.}
    \end{subfigure}
    \hfill
    \begin{subfigure}{0.22\linewidth}
        \includegraphics[width=\linewidth]{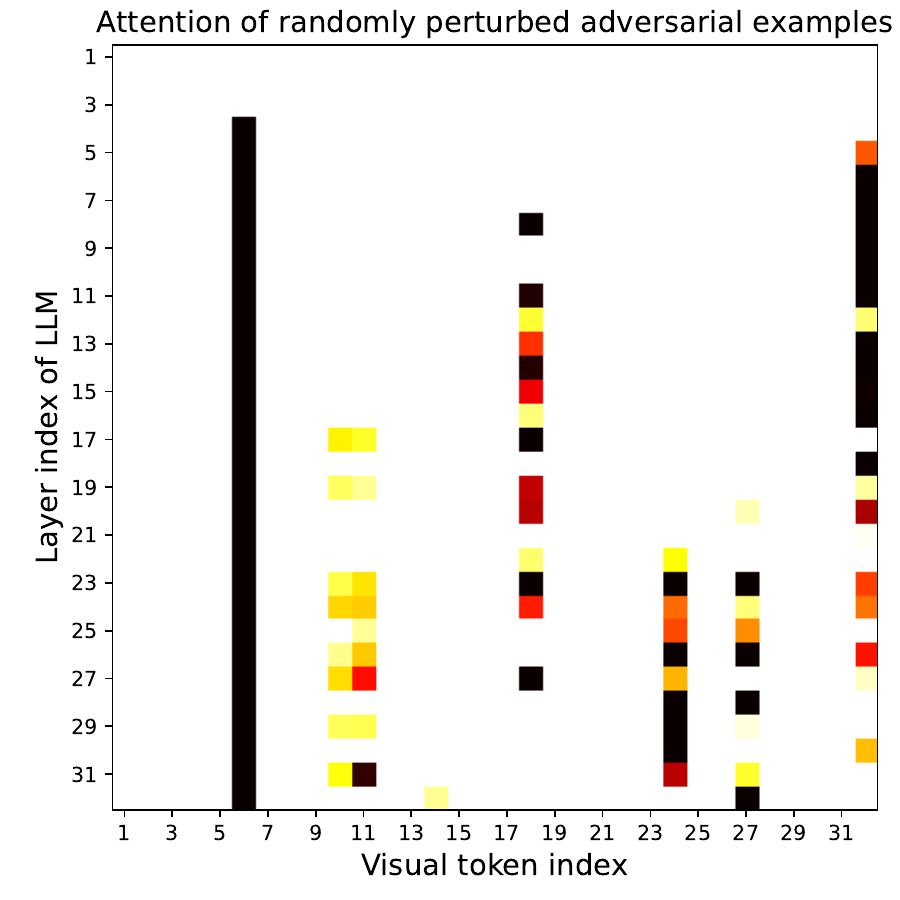}
        \caption{Reference attention $\mathbf{A}(x_i^R)$.}
    \end{subfigure}
    \hfill
    \begin{subfigure}{0.22\linewidth}
        \includegraphics[width=\linewidth]{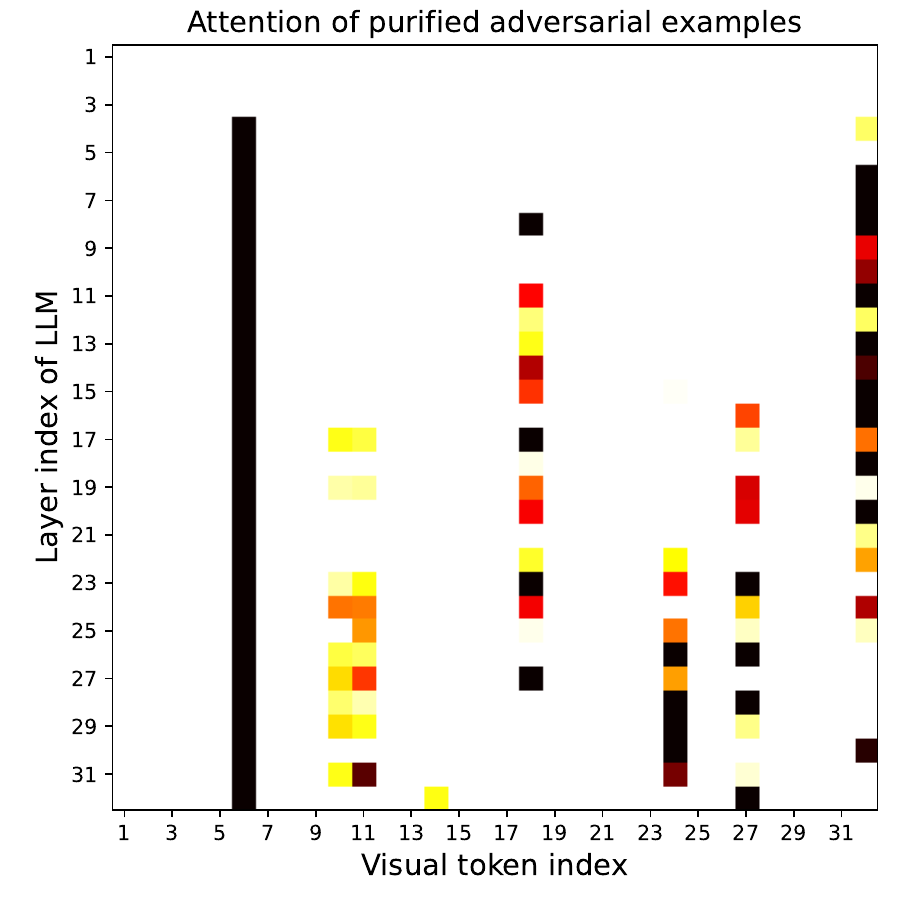}
        \caption{Purified attention $\mathbf{A}(x_i^p)$.}
    \end{subfigure}
    \caption{The visualization results of ``yes/no'' question. The question is ``Is this a toy train that a child could play with?''. The answers to the four images from (a) to (d) are ``no'', ``yes'', ``yes'', and ``no''.}
    \label{fig:attention_of_yesno}
\end{figure*}

\begin{figure*}[htbp]
    \centering
    \begin{subfigure}{0.22\linewidth}
        \includegraphics[width=\linewidth]{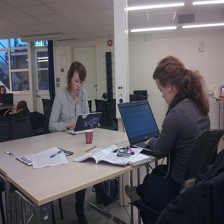}
        \caption{Clean image $x_i$.}
    \end{subfigure}
    \hfill
    \begin{subfigure}{0.22\linewidth}
        \includegraphics[width=\linewidth]{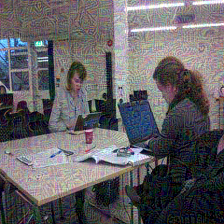}
        \caption{Adversarial image $x_i'$.}
    \end{subfigure}
    \hfill
    \begin{subfigure}{0.22\linewidth}
        \includegraphics[width=\linewidth]{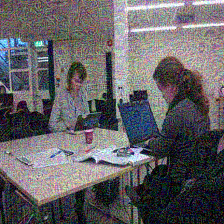}
        \caption{Perturbed image $x_i^R$.}
    \end{subfigure}
    \hfill
    \begin{subfigure}{0.22\linewidth}
        \includegraphics[width=\linewidth]{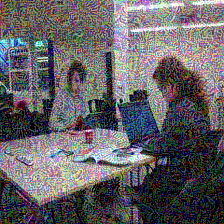}
        \caption{Purified image $x_i^p$.}
    \end{subfigure}
    \begin{subfigure}{0.22\linewidth}
        \includegraphics[width=\linewidth]{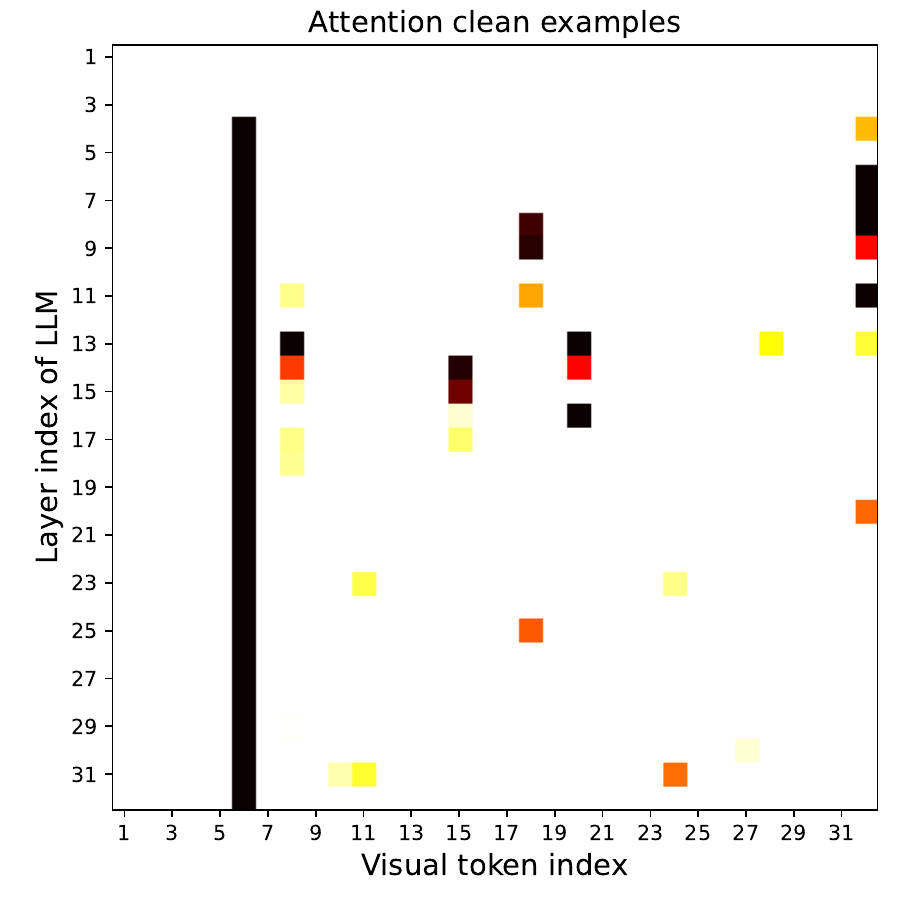}
        \caption{Clean attention $\mathbf{A}(x_i)$.}
    \end{subfigure}
    \hfill
    \begin{subfigure}{0.22\linewidth}
        \includegraphics[width=\linewidth]{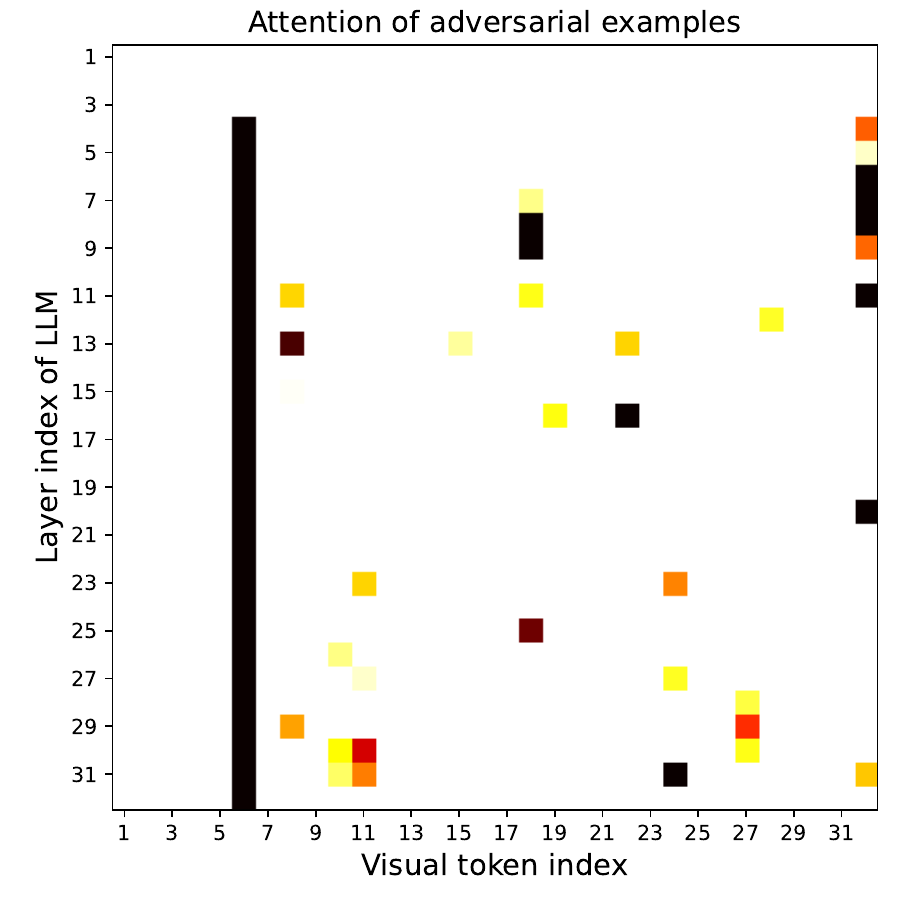}
        \caption{Adversarial attention $\mathbf{A}(x_i')$.}
    \end{subfigure}
    \hfill
    \begin{subfigure}{0.22\linewidth}
        \includegraphics[width=\linewidth]{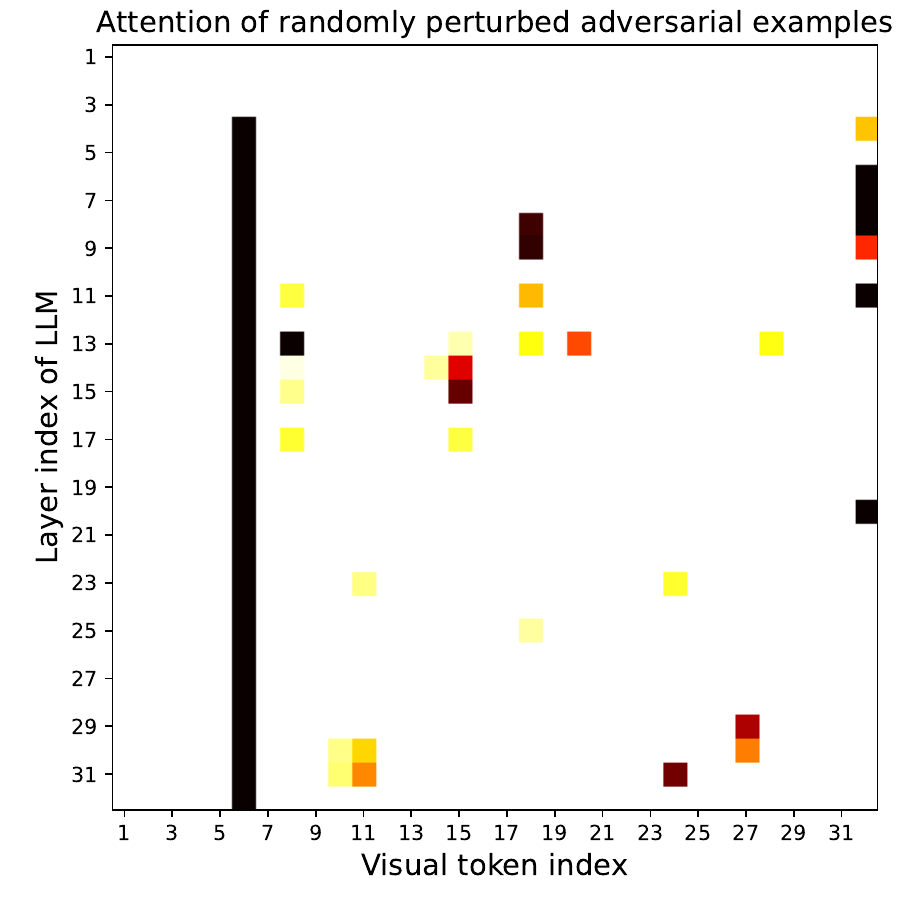}
        \caption{Reference attention $\mathbf{A}(x_i^R)$.}
    \end{subfigure}
    \hfill
    \begin{subfigure}{0.22\linewidth}
        \includegraphics[width=\linewidth]{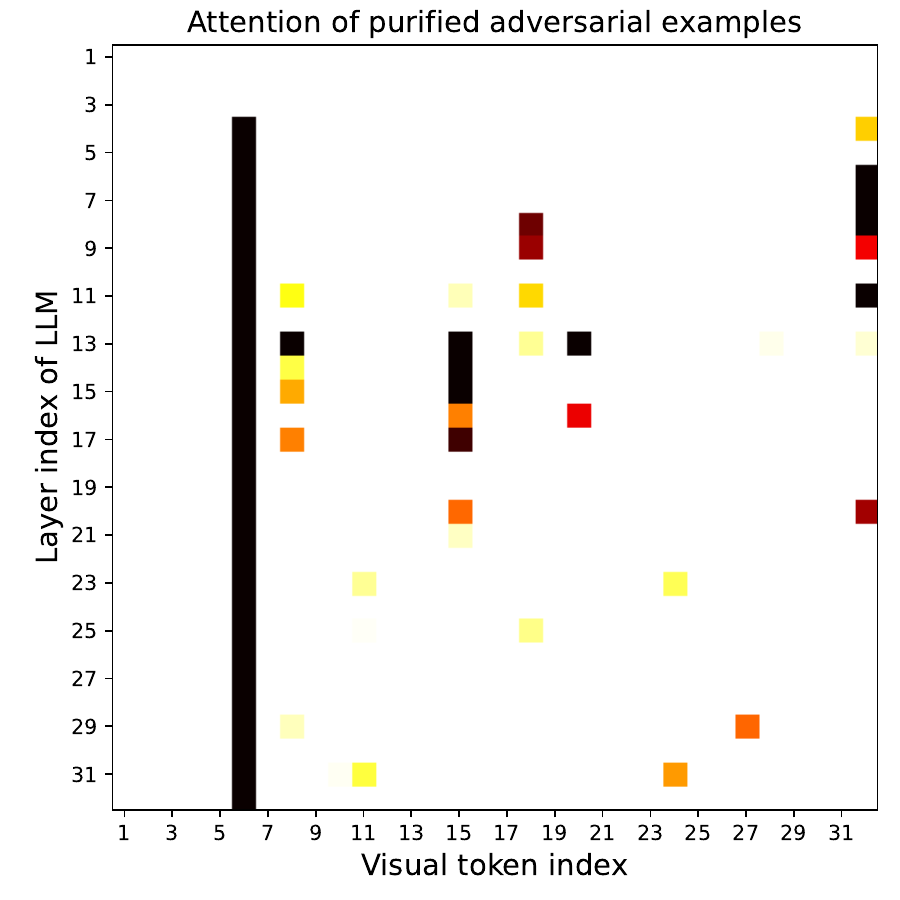}
        \caption{Purified attention $\mathbf{A}(x_i^p)$.}
    \end{subfigure}
    \caption{The visualization results of ``number'' question. The question is ``How many people are seated at this table?''. The answers to the four images from (a) to (d) are ``2'', ``0'', ``0'', and ``2''.}
    \label{fig:attention_of_number}
\end{figure*}

We present additional examples of clean images, adversarial images, randomly perturbed adversarial images, and purified images, along with their corresponding cross-modal attention. In \cref{fig:pipeline}, we have already demonstrated the results under C\&W attacks. Furthermore, we provide the results under PGD attacks in \cref{fig:attention_under_pgd}. The VQA questions are categorized into three types: ``yes/no'', ``number'', and ``other''. While we have previously shown the results for the ``other'' type of questions in \cref{fig:pipeline}, we also include the visualization results of C\&W attacks for different question types in \cref{fig:attention_of_yesno} (yes/no) and \cref{fig:attention_of_number} (number).

The attention visualizations in \cref{fig:attention_under_pgd,fig:attention_of_number,fig:attention_of_yesno}, along with the quantitative metrics presented in \cref{tab:similarity_of_attneion_of_three_sections}, collectively support our hypothesis. Specifically, the attention distribution of the randomly perturbed adversarial example $\mathbf{A}(x_i^R)$ is found to be more similar to that of the clean example $\mathbf{A}(x_i)$ compared to the attention of the original adversarial example $\mathbf{A}(x_i')$. Additionally, the attention distribution of the purified example  $\mathbf{A}(x_i^p)$ also shows greater alignment with the clean attention $\mathbf{A}(x_i)$. Notably, we observe a positive correlation: the closer the attention distribution of the purified image is to that of the clean image, the more effective the purification process proves to be. This observed relationship highlights the critical role of cross-modal attention mechanisms in the context of adversarial example purification.

\end{document}